\pdfoutput=1

\documentclass[11pt]{article}

\usepackage{setspace}
\usepackage{hyperref}
\usepackage[table,dvipsnames]{xcolor}
\usepackage[final]{acl}

\usepackage{times}
\usepackage{latexsym}

\usepackage[T1,T5]{fontenc}

\usepackage[utf8]{inputenc}
\usepackage{microtype}
\usepackage{inconsolata}
\usepackage{amssymb}
\usepackage{pifont}

\renewcommand{\encodingdefault}{T1}

\usepackage{graphicx}
\usepackage{tabularx}
\usepackage{amsmath}
\usepackage{booktabs}
\usepackage{multirow}
\usepackage{caption}
\usepackage{subcaption}
\usepackage{algorithm}
\usepackage{algpseudocode}
\usepackage{amsfonts}
\usepackage{verbatim}
\usepackage{tikzducks}
\usepackage{adjustbox}
\usepackage{arydshln}
\usepackage{rotating,booktabs}
\usepackage{makecell}
\usepackage{lipsum}

\usepackage{csquotes}
\usepackage{tablefootnote}

\usepackage{tikz}
\usetikzlibrary{shapes, arrows.meta, positioning}

\usepackage{scalerel,xparse}
\usepackage{fix-cm}

\usepackage{array}
\usepackage{xspace}
\usepackage{tikz}
\usetikzlibrary{matrix,shapes.arrows, positioning, arrows.meta}

\newcolumntype{L}[1]{>{\raggedright\let\newline\\\arraybackslash\hspace{0pt}}m{#1}}
\newcolumntype{C}[1]{>{\centering\let\newline\\\arraybackslash\hspace{0pt}}m{#1}}
\newcolumntype{R}[1]{>{\raggedleft\let\newline\\\arraybackslash\hspace{0pt}}m{#1}}

\makeatletter
\def\adl@drawiv#1#2#3{%
        \hskip.5\tabcolsep
        \xleaders#3{#2.5\@tempdimb #1{1}#2.5\@tempdimb}%
                #2\z@ plus1fil minus1fil\relax
        \hskip.5\tabcolsep}
\newcommand{\cdashlinelr}[1]{%
  \noalign{\vskip\aboverulesep
           \global\let\@dashdrawstore\adl@draw
           \global\let\adl@draw\adl@drawiv}
  \cdashline{#1}
  \noalign{\global\let\adl@draw\@dashdrawstore
           \vskip\belowrulesep}}
\makeatother

\makeatletter
\DeclareRobustCommand\onedot{\futurelet\@let@token\@onedot}
\def\@onedot{\ifx\@let@token.\else.\null\fi\xspace}

\makeatother

\usepackage{todonotes}
\usepackage{tabularx}
\usepackage{nicefrac}       
\usepackage{microtype}      
\usepackage{lineno}
\usepackage{amsmath}
\usepackage{amssymb}
\usepackage{amsfonts}
\usepackage{diagbox} 
\usepackage{float}
\usepackage{subcaption}
\usepackage{placeins}
\usepackage{multirow}
\usepackage{tabularx}

\usepackage{makecell}
\usepackage{placeins}
\usepackage{pifont}
\usepackage{array} 
\usepackage{ragged2e}
\newcolumntype{Y}{>{\centering\arraybackslash}X}
\newcolumntype{Z}{>{\raggedright\arraybackslash}X}
\usepackage{graphicx}
\usepackage{geometry}
\usepackage{setspace}
\usepackage{pgfplots}
\pgfplotsset{compat=1.17}

\usepackage{cleveref}
\usepackage{fontawesome5}

\makeatletter
\newcommand*\iftodonotes{\if@todonotes@disabled\expandafter\@secondoftwo\else\expandafter\@firstoftwo\fi}
\makeatother

\newcommand{\sparagraph}[1]{\vspace{0.0mm}\noindent\textbf{#1.}}

\definecolor{bluepigment}{rgb}{0.2, 0.2, 0.6}
\definecolor{ballblue}{rgb}{0.13, 0.67, 0.8}
\definecolor{bleudefrance}{rgb}{0.19, 0.55, 0.91}

\definecolor{spotifygreen}{RGB}{30, 215, 96}
\definecolor{twitterblue}{RGB}{29, 161, 242}

\definecolor{softblue}{RGB}{143, 178, 200}
\definecolor{softblue}{RGB}{132, 197, 250}
\definecolor{t5blue}{RGB}{207, 226, 243}
\definecolor{vibrantorange}{RGB}{240, 177, 110}
\definecolor{t5red}{RGB}{244, 204, 204}
\definecolor{freshgreen}{RGB}{111, 207, 151}
\definecolor{t5green}{RGB}{217, 234, 211}
\definecolor{t5yellow}{RGB}{255, 242, 204}

\newcommand{\rparagraph}[1]{\vspace{0.6mm}\noindent\textbf{#1.}}
\newcommand{\vcal}{\mathcal{V}}
\newcommand{\mcal}{\mathcal{M}}
\newcommand{\vcalmil}{\mathcal{V}_\text{1M}}
\newcommand{\vcalinit}{\mathcal{V}_\text{init}}
\newcommand{\vcalnew}{\mathcal{V}_\text{new}}
\newcommand{\vcallarge}{\mathcal{V}_\text{large}}

\usepackage[breakable]{tcolorbox}
\newtcolorbox{custombox}[1]{
  colback=green!5!white,   
  colframe=green!60!black, 
  boxrule=0.5mm,           
  arc=4mm,                 
  boxsep=1mm,              
  left=6mm,                
  right=6mm,               
  top=5mm,                 
  breakable=true,
  bottom=5mm,               
  title=#1
}
\usepackage{tabularx}
\usepackage{placeins}
\usepackage{pifont}

\definecolor{azure}{rgb}{0.0, 0.5, 1.0}
\definecolor{babyblue}{rgb}{0.54, 0.81, 0.94}

\usepackage{amsmath,amsfonts,bm}

\def\eqref#1{equation~\ref{#1}}

\def\1{\bm{1}}

\DeclareMathAlphabet{\mathsfit}{\encodingdefault}{\sfdefault}{m}{sl}
\SetMathAlphabet{\mathsfit}{bold}{\encodingdefault}{\sfdefault}{bx}{n}

\title{Retrofitting Large Language Models with Dynamic Tokenization}

\author{Darius Feher\thanks{Now at Google.} \quad Ivan Vulić \quad Benjamin Minixhofer\\
        University of Cambridge \\
        \texttt{feherdarius7@gmail.com, \{iv250, bm644\}@cam.ac.uk}
}

\begin{document}

\maketitle

\begin{abstract}
Current language models (LMs) use a fixed, \textit{static} subword tokenizer. This default choice typically results in degraded efficiency and language capabilities, especially in languages other than English.
To address this issue, we challenge the static design and propose retrofitting LMs with \textit{dynamic tokenization}: a way to dynamically decide on token boundaries based on the input text via a subword-merging algorithm inspired by byte-pair encoding. We merge frequent subword sequences in a batch, then apply a pre-trained embedding-prediction hypernetwork to compute the token embeddings on-the-fly. For encoder-style models (e.g., XLM-R),
this on average reduces token sequence lengths by >20\% across 14 languages while degrading performance by less than 2\%. The same method applied to prefilling and scoring in decoder-style models (e.g., Mistral-7B) results in minimal performance degradation at up to $17\%$ reduction in sequence length.
Overall, we find that dynamic tokenization can mitigate the limitations of static tokenization by substantially improving inference speed and promoting fairness across languages, enabling more equitable and adaptable LMs.
\end{abstract}

\section{Introduction}
\label{sec:introduction}
(Large) Language Models (LMs) are the backbone of modern NLP applications, enabling advanced language understanding and generation.
However, their effectiveness heavily relies on their tokenizers, which are responsible for \textit{tokenizing} the input~\citep{rust2020good, fujii2023different, toraman2023impact, ali2023tokenizer, minaee2024large, minixhofer2024zero}. This fundamental step involves breaking raw text into smaller units called \textit{tokens}, which are part of the tokenizer's \textit{vocabulary}. Since machines can only work with numerical data, tokens are converted into numerical IDs, which are then used to obtain \textit{embeddings} --- fixed-size vectors that serve as the model's representation of a token.

\begin{table}[h!]
\scriptsize
\centering
\begin{tabular}{@{}p{0.1\linewidth} p{0.65\linewidth}p{0.12\linewidth}@{}}
\toprule
\textbf{Language} & \textbf{Original Subword Tokenization} & \textbf{\#tokens} \\
\midrule
\textbf{English} & A sub\textcolor{red}{\textbf{/}}stantial im\textcolor{red}{\textbf{/}}prove\textcolor{red}{\textbf{/}}ment fosters further im\textcolor{red}{\textbf{/}}prove\textcolor{red}{\textbf{/}}ment\textcolor{red}{\textbf{/}}s & \multicolumn{1}{c}{12} \\ [6pt]
\textbf{Swahili} & U\textcolor{red}{\textbf{/}}bor\textcolor{red}{\textbf{/}}esh\textcolor{red}{\textbf{/}}aj\textcolor{red}{\textbf{/}}i mk\textcolor{red}{\textbf{/}}ub\textcolor{red}{\textbf{/}}wa una\textcolor{red}{\textbf{/}}ku\textcolor{red}{\textbf{/}}za u\textcolor{red}{\textbf{/}}bor\textcolor{red}{\textbf{/}}esh\textcolor{red}{\textbf{/}}aj\textcolor{red}{\textbf{/}}i za\textcolor{red}{\textbf{/}}idi  & \multicolumn{1}{c}{18} \\ [6pt]

\midrule
\textbf{\#merges} & \textbf{Dynamic Tokenization} & \textbf{\#tokens} \\
\midrule

\multicolumn{1}{c}{1} & A sub\textcolor{red}{\textbf{/}}stantial \textcolor{azure}{\it improve}\textcolor{red}{\textbf{/}}ment fosters further \textcolor{azure}{\it improve}\textcolor{red}{\textbf{/}}ment\textcolor{red}{\textbf{/}}s & \multicolumn{1}{c}{10 (83\%)} \\
\multicolumn{1}{c}{1} & U\textcolor{red}{\textbf{/}}\textcolor{azure}{\it boresh}\textcolor{red}{\textbf{/}}aj\textcolor{red}{\textbf{/}}i mk\textcolor{red}{\textbf{/}}ub\textcolor{red}{\textbf{/}}wa una\textcolor{red}{\textbf{/}}ku\textcolor{red}{\textbf{/}}za u\textcolor{red}{\textbf{/}}\textcolor{azure}{\it boresh}\textcolor{red}{\textbf{/}}aj\textcolor{red}{\textbf{/}}i za\textcolor{red}{\textbf{/}}idi  & \multicolumn{1}{c}{16 (89\%)} \\

\midrule

\multicolumn{1}{c}{2} & A sub\textcolor{red}{\textbf{/}}stantial \textcolor{azure}{\it improvement} fosters further \textcolor{azure}{\it improvement}\textcolor{red}{\textbf{/}}s & \multicolumn{1}{c}{8 (67\%)} \\

\multicolumn{1}{c}{2} & U\textcolor{red}{\textbf{/}}\textcolor{azure}{\it boreshaj}\textcolor{red}{\textbf{/}}i mk\textcolor{red}{\textbf{/}}ub\textcolor{red}{\textbf{/}}wa una\textcolor{red}{\textbf{/}}ku\textcolor{red}{\textbf{/}}za u\textcolor{red}{\textbf{/}}\textcolor{azure}{\it boreshaj}\textcolor{red}{\textbf{/}}i za\textcolor{red}{\textbf{/}}idi  & \multicolumn{1}{c}{14 (78\%)} \\

\midrule

\multicolumn{1}{c}{4} & A \textcolor{azure}{\it substantial} \textcolor{azure}{\it improvement} fosters further \textcolor{azure}{\it improvements} &\multicolumn{1}{c}{\textcolor{ForestGreen}{6 (50\%)}} \\ 

\multicolumn{1}{c}{11} & \textcolor{azure}{\it Uboreshaji mkubwa unakuza uboreshaji zaidi} & \multicolumn{1}{c}{\textcolor{ForestGreen}{5 (28\%)}} \\ 

\bottomrule
\end{tabular}
\vspace{-0.5mm}
\caption{Comparison of static subword vs dynamic tokenization for the same sentences in \textbf{English} and \textbf{Swahili}.  Embeddings for tokens in {\it \textcolor{azure}{blue}} are obtained using a hypernetwork (HN) which composes the subword-level embeddings, as highlighted by \textcolor{red}{\textbf{/}}. The last row shows the number of merges required to achieve word-level tokenization, serving as a `compression upper bound' of the proposed approach. The percentages show the fraction of the original token count remaining after merging.}
\label{tab:example_dynamic_tokenization}
\vspace{-2mm}
\end{table}

The majority of contemporary LMs rely on \textit{subword tokenizers} (e.g., \citeauthor{devlin2018bert}, \citeyear{devlin2018bert}; \citeauthor{touvron2023llama}, \citeyear{touvron2023llama}) that are inherently static, as their vocabularies remain fixed post-training. This rigidity limits the model's adaptability, requiring expensive retraining to update both the vocabulary and embeddings~\citep{dagan2024getting}. Moreover, subword tokenizers struggle with handling sequences of numbers \citep{golkar2023xval}, are sensitive to spelling errors \citep{sun2020adv, xue2022byt5} and often suffer from over-segmentation in languages other than English \citep{wang2021multi}. This leads to inequitable performance across languages, increasing inference costs, latency, and reducing overall model effectiveness \citep{ahia2023all}. While character- or byte-level tokenization provides a potential solution, it produces long token sequences which brings additional challenges, such as the need for dynamic compute allocation or token pooling to stay competitive to models using subword tokenization in terms of efficiency~\citep{nawrot2022efficient}.
These issues underscore the need for a more flexible or dynamic tokenization that adapts token boundaries based on the input text. This is the focus of our work. Specifically, we introduce a way of \textit{retrofitting} existing (subword-based) LMs with dynamic tokenization.

Our proposed dynamic tokenization approach focuses on improving \textit{efficiency} and \textit{cross-lingual fairness} by repurposing a hypernetwork (HN) introduced by \citet{minixhofer2024zero} --- originally intended for zero-shot transfer across tokenizers --- to support dynamic tokenization; see Table~\ref{tab:example_dynamic_tokenization} for an illustrative example, and later Figure~\ref{fig:dynamic_tokenization_framework}. This adaptation uses the HN to dynamically generate token embeddings on-the-fly, substantially reducing token sequence lengths at minimal performance degradation, and also effectively enabling an \textit{unbounded vocabulary} for encoding text.

This approach, as our extensive experiments demonstrate, is highly beneficial for \textit{prefilling} (computing the key-value states of a prompt) and \textit{scoring} (computing the likelihood of a text) with generative models. However, applying it to autoregressive next-token generation is more challenging since softmax normalization over an unbounded vocabulary is intractable. We thus aim to achieve the benefits of dynamic tokenization for autoregressive generation by expanding to a large, but still bounded vocabulary (in practice, 1M tokens); we introduce a highly efficient method to deal with the large vocabulary. It is based on an approximate nearest neighbor index to overcome the parameter overhead and the softmax bottleneck~\citep{yang2018breaking} by dynamically retrieving tokens.



\rparagraph{Contributions} \textbf{1)} We propose an approach of retrofitting LMs with dynamic tokenization, achieving a 22.5\% reduction in token sequence length on XNLI and a 26.4\% reduction on UNER, with minimal performance degradation. This improves inference speed and leads to fairer compute allocation across languages (see \Cref{sec:results}). \textbf{2)} We adapt the same method to \textit{prefilling} and \textit{scoring} in decoder-style LLMs, achieving minimal performance degradation at up to 17\% sequence length reduction. \textbf{3)} Since naïvely applying our method to autoregressive generation is intractable, we further investigate generation with a large but bounded vocabulary of 1M tokens, achieving additional gains in efficiency. Our code is publicly available at  \hspace{4mm} {\faGithub \hspace{0.3mm}} \href{https://github.com/DariusFeher/dynamic-tokenization}{github.com/DariusFeher/dynamic-tokenization}.




\section{Background and Related Work}
\label{sec:related}
\rparagraph{Tokenizers: Preliminaries} We follow the tokenizer definition used by \citet{uzan2024greed} and \citet{minixhofer2024zero}. Let $\vcal$ denote a \textit{vocabulary}, and $T$ a \textit{tokenization function}. A tokenizer is then a tuple consisting of these two components, $(\vcal, T)$. The vocabulary $\vcal$ contains the set of tokens, while the tokenization function $T$ is used to segment the input text into smaller units, which are part of $\vcal$. Importantly, for a given $\vcal$, there are multiple ways to encode the same input text into a sequence of tokens~\citep{hofmann2022embarrassingly}, with $T$ determining the specific encoding method. After tokenizing the input text into a sequence of tokens, each token is then mapped to a continuous vector representation using the embedding function $E_\phi: \vcal \to \mathbb{R}^{d_{\text{model}}}$, parameterized by a matrix $\phi$. This matrix serves as a lookup table, assigning each token a unique $d_{\text{model}}$-dimensional vector.

\rparagraph{Static Tokenizers} Existing tokenizers implement character-, byte-, subword- and word-level tokenization. Character-~\citep{boukkouri2020characterbert, tay2021charformer, clark2022canine} and byte-level~\citep{xue2022byt5, yu2024megabyte} methods offer advantages such as small vocabularies and increased robustness to noise, helping in handling rare words and low-resource languages. However, they suffer from reduced processing speed due to longer token sequences or required sequence pooling, impacting training and inference efficiency \citep{clark2022canine, nawrot2022efficient}. Furthermore, byte-level tokenizers are biased against non-Latin scripts~\citep{limisiewicz-etal-2024-myte}. 

Word tokenization methods provide faster processing with shorter token sequences, but struggle with out-of-vocabulary (OOV) words and require large vocabularies. A commonly used `middle ground' is thus subword tokenization, which breaks down the text into smaller, more manageable units, such as pieces of words or entire words.
Techniques like Byte-Pair Encoding \citep[BPE;][]{sennrich2015neural}, WordPiece~\citep{schuster2012japanese} and UnigramLM~\citep{kudo2018subword}, handle OOV words by breaking them into known subword units, while also maintaining manageable vocabulary sizes and sequence lengths.
Crucially, all these methods are \textit{static}, relying on a predefined vocabulary $\vcal$ that does not adapt to new data post-training, limiting adaptability to new words or evolving language. This is problematic especially in multilingual contexts, leading to over-segmentation, reduced performance, and increased inference costs in languages other than English~\citep{ahia2023all}. Moreover, BPE has been shown to be suboptimal for LM pretraining, limiting downstream performance \citep{bostrom2020byte}. 
These issues highlight the need for dynamic tokenization to potentially achieve higher efficiency and more equitable performance across languages.



\rparagraph{Vocabulary Expansion} Previous work on adaptive tokenization focused on expanding vocabularies with domain- or language-specific tokens. However, this greatly increases the size of the embedding matrix --- sometimes accounting for up to 93\% of model parameters~\citep{liang2023xlm} --- which limits how many new tokens can be effectively added and results in inefficient parameter allocation. New token embeddings are typically initialized with heuristics~\citep{minixhofer2022wechsel, gee2024fast, liu2024ofa, gee2024fast} and require additional training for optimal performance, restricting real-time adaptation. 
We use Fast Vocabulary Transfer~\citep[FVT;][]{gee2024fast} as a baseline heuristic. FVT generates embeddings for a new token by tokenizing it with the original tokenizer and averaging the embeddings of its subword tokens. Alternative multi-token generation techniques like Copy-Generator~\citep{lan2023copyneed} and Nearest Neighbor Speculative Decoding~\citep{li2024nearest} use token databases and nearest neighbor retrieval, but face challenges with factual accuracy and computational efficiency. In contrast, our pre-trained HN efficiently generates individual token embeddings removing the need for fine-tuning across domains, addressing both the parameter overhead of vocabulary expansion and the computational requirements of multi-token generation.

\rparagraph{Token Embedding Prediction} Instead of relying on heuristics to initialize the embeddings of new tokens, more advanced methods predict them using neural networks. This includes using neural networks to predict the embeddings of rare~\citep{schick2019attentive} or OOV~\citep{pinter2017mimicking} words in traditional word models, an approach later adapted by~\citet{schick2020bertram} for \textsc{BERT}~\citep{devlin2018bert}. However, these methods are limited to expanding the existing tokenizers rather than enabling transfer to an entirely different tokenizer.
In contrast, Zero-Shot Tokenizer Transfer~\citep[ZeTT;][]{minixhofer2024zero} enables transferring LMs to any arbitrary, but \textit{fixed/static} tokenizer. This extends beyond only enabling vocabulary extension to full transfer to a completely new tokenizer while preserving the LM's performance to a large extent in most cases by using a hypernetwork to predict the token embeddings. Our key insight is that these hypernetworks can be repurposed to enable dynamic tokenization.\footnote{On the other side, this makes our approach depend on the availability of pretrained hypernetworks for the target model. We discuss this constraint in Appendix~\ref{sec:hypernetwork_training}.}


\section{Methodology}
\label{sec:method}
\sparagraph{Problem Formulation}
\textit{Dynamic tokenization} changes the traditional static
encoding process by adaptively adjusting token boundaries based on the input text, continuously updating the vocabulary $\vcal$ and the tokenization function $T$.
This contrasts with the static tokenization, where $\vcal$ and $T$ remain fixed post-training. 
%
More formally, let the initial tokenizer be $(\vcal_\text{init}, T_\text{init})$. As the LM operates with new text data $\mathcal{D}$, the tokenization function $T_\text{init}$ is updated to $T_\text{new}$. The update process can be represented by the function $\mathcal{U}$:
\begin{equation}
    T_\text{new}(\mathcal{D}) = \mathcal{U}(T_\text{init}(\mathcal{D}))
\end{equation}
To retrofit an LM pre-trained with subword tokenization to dynamic tokenization, two steps are required: \emph{(1)} deciding on a tokenization $T_\text{new}$; and \emph{(2)} obtaining the token embeddings. This approach can be applied to any case where the (subword-level) token sequence is known in advance. 

\subsection{Dynamic Tokenization via BPE-Style Compression}\label{sec:dynamic_tokenization_bpe}

\sparagraph{Deciding on a Dynamic Tokenization}
Let $\mathcal{D}$ represent the input data to be tokenized. The first step in dynamic tokenization involves updating the initial tokenization $T_\text{init}$ to a new function $T_\text{new}$, using the update function $\mathcal{U}$. Since our focus is on efficiency, this update aims to minimize over-segmentation in the input data $\mathcal{D}$, resulting in a more compact representation for $\mathcal{D}$ (i.e., $|T_\text{new}\left(\mathcal{D}\right)| \leq |T_\text{init}\left(\mathcal{D}\right)|$).


Importantly, given that 
LMs operate at \textit{batch-level}, $\mathcal{U}$ is specifically applied at this level on $\mathcal{D}_\text{batch}$. This allows $\mathcal{U}$ to dynamically adapt the tokenization to the unique linguistic features in each batch.

To define $\mathcal{U}\left(T_\text{init}(\mathcal{D}_\text{batch})\right)$, we take inspiration from BPE~\citep{sennrich2015neural}.
Specifically, for each batch $\mathcal{D}_\text{batch}$ tokenized under the initial scheme $T_{\text{init}}$, we begin with a \textit{batch-specific vocabulary} $\vcal_{\text{new}}$ comprised of all unique subword tokens present in $T_\text{init}\left(\mathcal{D}_\text{batch}\right)$. We then perform a fixed number of merge operations, $m$, combining the most frequent adjacent tokens within the batch, continuously refining the tokenization to better compress $\mathcal{D}_\text{batch}$.

We formally define the update function $\mathcal{U}$ as:
\begin{equation}
\begin{split}
    \mathcal{U}: \left(T_\text{init}\left(\mathcal{D}\right), m\right) \to T_\text{new}\left(\mathcal{D}\right) \\
    \text{ with } \quad |T_\text{new}\left(\mathcal{D}\right)| \leq |T_\text{init}\left(\mathcal{D}\right)|,
\end{split}
\end{equation}
\noindent where $m$ represents the number of merge operations to perform. Since the BPE-style merging process is applied to $T_\text{init}\left(\mathcal{D}_\text{batch}\right)$ --- the data we wish to tokenize --- we implicitly tokenize this batch under the new tokenization scheme $T_\text{new}\left(\mathcal{D}_\text{batch}\right)$ by sequentially applying merge operations. This allows us to simplify the training and tokenization processes of traditional BPE into a single, unified algorithm, outlined in Appendix~\ref{sec:bpe_dynamic_tokenization_lm}.\footnote{Additionally, since the new tokenization is only applied to the specific batch data $\mathcal{D}_\text{batch}$ and not reused for other text, there is no need to store or compute new merge rules $\mcal_\text{new}$ or maintain an expanded vocabulary $\vcal_\text{new}$.}

\begin{figure}[t!] 
\centering    
\includegraphics[width=0.47\textwidth, trim={40pt 2pt 5pt 10pt}, clip]{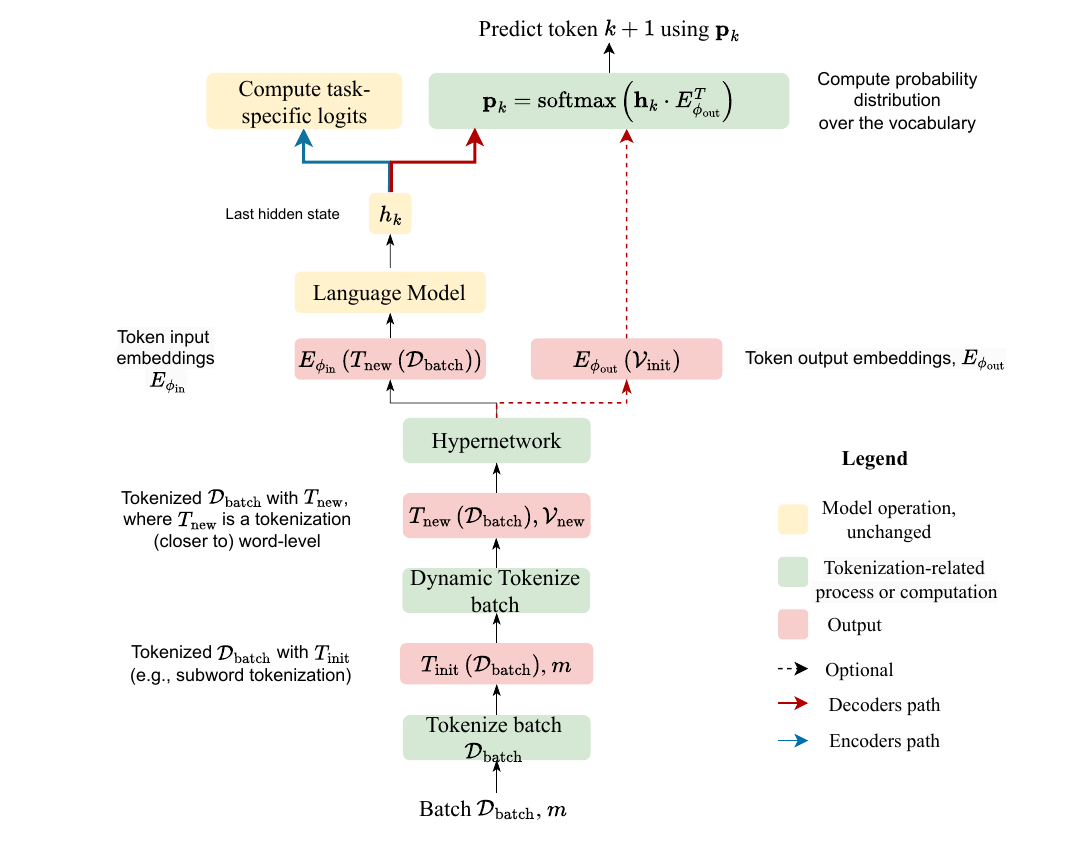}
\vspace{-1.5mm}
\caption{Dynamic tokenization applied to encoders and decoders LMs.}
\label{fig:dynamic_tokenization_framework}
\vspace{-2mm}
\end{figure}

Subword-level tokenization represents the starting point or the \textit{lower-bound} for the new tokenization, $T_\text{new}$, obtained when $m=0$, and equivalent with $T_\text{init}$.
On the other hand, we consider word-level, or, more precisely, pre-token-level,\footnote{Pre-tokens are preliminary units often equivalent to words~\citep{mielke2021between}. For simplicity of terminology, we use \textit{pre-tokens} and \textit{words} interchangeably, but note that `pre-token' is the more precise term.} as the \textit{upper-bound} for $T_\text{new}$. In other words, we constrain the merging process to never merge adjacent tokens which are part of different words.

\rparagraph{Obtaining Token Embeddings}
After mapping the tokens from the initial tokenization to a more compact tokenization, $T_\text{init}\left(\mathcal{D}_\text{batch}\right) \to T_\text{new}\left(\mathcal{D}_\text{batch}\right)$, we need to obtain the embeddings for all tokens $t \in T_\text{new}\left(\mathcal{D}_\text{batch}\right)$. To achieve this, we repurpose the HN trained by \citet{minixhofer2024zero}. While the HN was originally intended to transfer an LM to a fixed, static tokenizer, we observe that it can also be used to achieve dynamic tokenization: since the HN \textit{amortizes} over the tokenization function (i.e., embedding predictions for every token are independent of each other), it does not require a static (or even bounded) vocabulary.
Therefore, for each $ t \in T_\text{new}\left(\mathcal{D}_\text{batch}\right) $, we apply the hypernetwork $ H_\theta $ to obtain its embedding:
\begin{equation}\label{eq:eq_hn_apply_to_vnew}
 E_{\phi_\text{new}}\left(t\right) = H_\theta \left(t\right), \quad \forall t \in T_\text{new}\left(\mathcal{D}_\text{batch}\right)
\end{equation}
where $E_{\phi_\text{new}}\left(t\right)$ is the embedding for token $t$ and $\phi_\text{new}$ is the matrix corresponding to the batch-specific vocabulary, $\vcal_\text{new}$.
This process can alternatively also be viewed as transferring the LM to a new tokenizer $\left(\vcal_\text{new}, T_\text{new}\right)$ for each batch, dynamically adjusting token boundaries based on the specific data within that batch.
Recall that it is applicable to any case where the token sequence is known in advance, i.e., any use-case of encoder-style LMs, as well as prefilling and scoring of generative (decoder-style) LMs, as shown in Figure~\ref{fig:dynamic_tokenization_framework}.

\section{Experimental Setup}
\label{sec:setup}
\sparagraph{Models} We use \textsc{XLM-R}~\citep{conneau2020unsupervised} as the representative multilingual encoder-style LM.
To test our method on a decoder-style model LM, we use both the base and instruct versions of Mistral-7B~\citep{jiang2023mistral}. This choice is partially due to the fact that the two established models come with pre-trained HNs.

\rparagraph{Datasets} For our XLM-R experiments, we use two datasets: Cross-lingual Natural Language Inference~\citep[XNLI;][]{conneau2018xnli}, and Universal Named Entity Recognition~\citep[UNER;][]{mayhew2023universal}. These datasets quantify the effect of dynamic tokenization across a total of 14 languages, with XNLI focusing on \textit{sentence-level} and UNER on \textit{token-level} understanding.\footnote{For XNLI, we evaluate on 13 different languages: Arabic, Bulgarian, German, Greek, English, Spanish, French, Hindi, Russian, Swahili, Turkish, Urdu, Vietnamese. Similarly, for UNER, we train our adapters on English, ``en\_ewt'' training split, and evaluate on 4 languages: English, German, Portuguese, and Russian.} For Mistral-7B experiments, we use the following evaluation benchmarks: the ``lite'' version of Global-MMLU~\citep{singh2024global} in English, French, German, Spanish and Portuguese, and the English Multi-Turn Benchmark~\citep[MT-Bench;][]{chiang2023can}.



\rparagraph{Embeddings} 
We compare the performance of the model using (i) the original embeddings, (ii) FVT embeddings\footnote{We use FVT since it achieves comparable performance to FOCUS \citep{dobler2023focus} while being substantially faster than FOCUS~\citep{minixhofer2024zero}, which is crucial for our dynamic setup.} (see Section~\ref{sec:related}) and (iii) HN-generated embeddings.

\rparagraph{Hyperparameters} Appendix~\ref{sec:reproducibility_details} details the hyperparameter settings used in our experiments.

\subsection{Experiments with Encoder Models}\label{sec:encoder_experiments}


We train a LoRA adapter~\citep{hu2021lora} for both \textit{task} --- natural language inference for XNLI and named entity recognition for UNER --- and \textit{dynamic tokenization} adaptation. The adapter jointly learns to adapt to the task and operate with coarser token granularities. We perform two experiments: \emph{(1)} training an adapter with a fixed number of merges $m$ and \emph{(2)} training an adapter with $m$ sampled from a Uniform distribution.\footnote{Our preliminary experiments included selecting $m$ from a Gaussian distribution. However, this yielded suboptimal results. Additionally, we investigated disentangling task adaptation from tokenization adaptation, but this also led to suboptimal results; future work could re-investigate whether disentangling the task and the tokenization is possible.}

More precisely, for each batch, we first apply the static BPE tokenizer to obtain an initial subword-level token sequence. Next, we choose a merge count $m$ (either fixed or sampled), and perform dynamic tokenization to merge frequenct adjacent tokens. These merged tokens are embedded on-the-fly using the HN and passed to the frozen encoder augmented with LoRA. We then compute the loss and update only the LoRA weights.

\rparagraph{(1) Predetermined Number of Merges} 
Here, we train an adapter with dynamic tokenization that reduces sequence length by a fixed percentage of the maximum possible reduction. We set this percentage to $50\%$ for XNLI and $75\%$ for UNER. Note that $100\%$ reduction corresponds to the difference between the initial sequence length --- obtained when tokenizing with $T_\text{init}$ --- and the sequence length obtained with word-level tokenization (i.e., the number of words in the sequence). We apply the function $\mathcal{U}(T_\text{init}(\mathcal{D}_\text{batch}), m)$ for each batch to meet the specific reduction percentage in sequence length.
This is necessary since the correlation between the number of merges $m$ and the sequence length reduction varies across languages and datasets; for instance, 140 merges achieve $100\%$ relative sequence reduction on English XNLI, whereas 250 merges are required for the same reduction on Turkish XNLI.


\rparagraph{(2) Sampling from a Uniform Distribution} In the second approach, instead of using a fixed number of merges $m$ and applying the function $\mathcal{U}(T_\text{init}(\mathcal{D}_\text{batch}), m)$ with the same $m$ across the training batch, we introduce stochasticity into the tokenization process. Specifically, we explore the impact of sampling different numbers of merges from a Uniform distribution. By training the adapter with tokenizations sampled from this distribution, we hypothesize that the model will learn to be more robust to the type of dynamic tokenization used (i.e., the value of $m$). We sample a tokenizer per batch (i.e., a fixed $m$) rather than a tokenizer for each sample in the batch due to the high computational requirements of the latter. The tokenization function applied during training is then:
\begin{equation}
    \mathcal{U}(T_\text{init}(\mathcal{D}_\text{batch}), m),  \quad  m \sim \mathrm{U}(0, m_{\text{max}})
\end{equation}

\newcolumntype{R}{>{\raggedleft\arraybackslash}X}
\begin{table*}[t!h] 
\centering
\def\arraystretch{0.93}
\footnotesize
\setlength\tabcolsep{1.75pt} 
\begin{tabularx}{\linewidth}{llccccccccccccc|R}
\toprule
\multirow{2}{*}{\makecell{\textbf{Tokenization \&}\\\textbf{Embeddings}}} & \multirow{2}{*}{\makecell{\textbf{Adapter}}} & \multicolumn{13}{c}{\makecell{\textbf{Accuracy per Language (\%)}}}\\
\cmidrule(r){3-15}
 & & \textbf{ar} & \textbf{bg} & \textbf{de} & \textbf{el} & \textbf{en} & \textbf{es} & \textbf{fr} & \textbf{hi} & \textbf{ru} & \textbf{sw} & \textbf{tr} & \textbf{ur} & \textbf{vi} & \textbf{Avg.} \\
\midrule
\rowcolor{gray!20} (1) original & task & 71.6 & \textbf{76.5 }& \textbf{76.9} & 75.1 & \textbf{84.8} & 78.0 & \textbf{78.5} & 68.7 & 74.9 & \textbf{63.2} &\textbf{72.4} & 65.4 & \textbf{73.9} & 73.9 \\
(2) original, HN & task & \textbf{71.8} & \textbf{76.5} & 76.7 & \textbf{75.7} & 84.1 & \textbf{79.0} & 78.2 & \textbf{69.6} & \textbf{75.7} & 61.7 & 72.1 & \textbf{65.9} & 73.7 & \textbf{74.0} \\
\midrule
(3) word, HN & task & 67.1 & 72.8 & \textbf{74.9} & 71.5 & 82.5 & 77.1 & 75.6 & 66.2 & 72.0 & 59.2 & 67.4 & 64.9 & 73.4 & 71.1 \\
(4) word, FVT & task &  64.5 & 68.9 & 70.8 & 68.3 & 79.7 & 74.2 & 71.0 & 65.2 & 68.6 & 54.8 & 63.3 & 63.8 & 73.6 & 68.2 \\
(5) word, HN & task + $m_\text{50\%}$ & \textbf{67.8} & \textbf{74.2} & 74.3 & \textbf{72.4} & 83.2 & \textbf{78.3} & 75.7 & \textbf{66.6} & \textbf{72.9} & \textbf{61.3} & \textbf{67.5} & \textbf{66.4} & \textbf{75.0} & \textbf{72.0} \\
(6) word, HN & task + $m_\text{sampled}$ & 66.5 & 74.1 & 74.5 & 71.6 & \textbf{84.3} & 77.0 & \textbf{75.9} & 64.9 & 72.7 & 58.8 & 66.5 & 65.1 & 73.7 & 71.2\\
\midrule
$\Delta_\text{Acc. (\%)}$ (1), (5) & & \textcolor{BrickRed}{-3.8} & \textcolor{BrickRed}{-2.3} & \textcolor{BrickRed}{-2.6} & \textcolor{BrickRed}{-2.7} & \textcolor{BrickRed}{-1.6} & \textcolor{JungleGreen}{0.3} & \textcolor{BrickRed}{-2.8} & \textcolor{BrickRed}{-2.1} & \textcolor{BrickRed}{-2.0} & \textcolor{BrickRed}{-1.9} & \textcolor{BrickRed}{-4.9} & \textcolor{JungleGreen}{1.0} & \textcolor{JungleGreen}{1.1} & \textcolor{BrickRed}{-1.9} \\
\multicolumn{2}{l}{$\Delta_\text{Length (\%)}$ original (1, 2), word (3-6)} & \textcolor{JungleGreen}{-31.4} & \textcolor{JungleGreen}{-25.1} & \textcolor{JungleGreen}{-22.8} & \textcolor{JungleGreen}{-33.2} & \textcolor{JungleGreen}{-14.7} & \textcolor{JungleGreen}{-17.3} & \textcolor{JungleGreen}{-17.3} & \textcolor{JungleGreen}{-21.8} & \textcolor{JungleGreen}{-28.2} & \textcolor{JungleGreen}{-28.4} & \textcolor{JungleGreen}{-29.4} & \textcolor{JungleGreen}{-17.5} & \textcolor{JungleGreen}{-5.9} & \textcolor{JungleGreen}{-22.5}\\
\bottomrule
\end{tabularx}
\vspace{-0.5mm}
\caption{Accuracy on \textbf{XNLI} \texttt{validation} with different adapters, tokenizations and embeddings. $\Delta_\text{Acc. (\%)}$ is the absolute accuracy change between word-level tokenization with the optimal adapter and HN embeddings (5) and the baseline (1) which uses original tokenization and embeddings. $\Delta_\text{Length. (\%)}$ represents the average decrease in token sequence length of word-level tokenization over the original. Boldface indicates the best result for a language when using \textit{original} subword-level tokenization or \textit{word-level} tokenization.}
\label{table:xnli_task_adapter_original}
\vspace{-1.5mm}
\end{table*}

\begin{table*}[t!] 
\centering
\def\arraystretch{0.95}
\footnotesize
\setlength\tabcolsep{3.75pt} 
\begin{tabularx}{.8\linewidth}{llccccc|R}
\toprule
\multirow{2}{*}{\makecell{\textbf{Tokenization \&}\\\textbf{Embeddings}}} & \multirow{2}{*}{\makecell{\textbf{Adapter}}} & \multicolumn{5}{c}{\makecell{\textbf{Language F1-score (\%)}}}\\
\cmidrule(r){3-7}
 & & \textbf{en\_ewt} & \textbf{de\_pud} & \textbf{pt\_bosque} & \textbf{pt\_pud} & \textbf{ru\_pud} & \textbf{Avg.} \\
\midrule
\rowcolor{gray!20} (1) original & task & \textbf{81.6} & 78.0 & \textbf{82.3} & \textbf{82.9} & \textbf{69.0} &  \textbf{78.8} \\
(2) original, HN & task & 80.9 & \textbf{78.3} & 80.8 & 82.3 & 68.4 & 78.1 \\
\midrule
(3) word, HN & task & 77.0 & 75.8 & 77.6 & 77.3 & 65.5 &  74.6\\
(4) word, FVT & task & 67.2 & 57.0 & 58.0 & 58.4 & 40.7 & 56.3 \\
(5) word, HN & task + $m_\text{75\%}$ & 80.5 & 75.0 & \textbf{80.5} & \textbf{81.3} & \textbf{67.9} & \textbf{77.0}  \\
(6) word, HN & task + $m_\text{sampled}$ & \textbf{81.3} & \textbf{76.3} & 78.5 & 80.2 & 67.1 & 76.7 \\
\midrule
$\Delta_\text{F1-score (\%)}$ (1), (5) & & \textcolor{BrickRed}{-1.1} & \textcolor{BrickRed}{-3.0} & \textcolor{BrickRed}{-1.8} & \textcolor{BrickRed}{-1.6} & \textcolor{BrickRed}{-1.1} & \textcolor{BrickRed}{-1.7}\\
\multicolumn{2}{l}{$\Delta_\text{Length (\%)}$ original (1, 2), word (3-6)}  &\textcolor{JungleGreen}{-17.6} & \textcolor{JungleGreen}{-30.5} & \textcolor{JungleGreen}{-24.1} & \textcolor{JungleGreen}{-24.2} & \textcolor{JungleGreen}{-35.8} & \textcolor{JungleGreen}{-26.4} \\
\bottomrule
\end{tabularx}
\vspace{-1mm}
\caption{F1-score on \textbf{UNER} with different adapters, tokenizations and embeddings. The results reported are on the \texttt{validation} split for \texttt{ewt} and \texttt{bosque} datasets, and \texttt{test} split for \texttt{pud} due to the availability.}

\label{table:uner_task_adapter_original}
\vspace{-2mm}
\end{table*}

\noindent where $m_\text{max}$ is determined  by $\mathcal{D}$ and represents the merge level yielding word-level tokenization.

\subsection{Experiments with Decoder Models}
\label{sec:decoder_experiments}
Unlike XLM-R, we do not train the Mistral-7B decoder model as we evaluate our method out-of-the-box on a pretrained checkpoint. We apply dynamic tokenization with a fixed number of merges $m$ to the input batch $\mathcal{D}_\text{batch}$. We evaluate performance trends across all sequence reduction percentages, 0\% to 100\%, where again 100\% corresponds to the reduction achieved with word-level tokenization. For prefilling, we compute the key-value states of the dynamically tokenized input sequence. For scoring, our goal is to compute the conditional probability $p(\text{suffix}|\text{prefix})$. Our key insight enabling dynamic tokenization is that we do not need to compute $p(\text{prefix})$ to compute the normalized conditional probability of some suffix relative to other suffixes. This means we can dynamically tokenize the prefix, then use the original tokenization to tokenize (and evaluate the probability of) the suffix. In practice, e.g., for MMLU this means processing each input prompt using dynamic tokenization only keeping the last hidden state $\mathbf{h}$, and compute a probability distribution over the four answer choices using $\mathbf{h}$ with the embeddings of the answer choices \texttt{A}, \texttt{B}, \texttt{C} and \texttt{D}. This setup also allows evaluating the quality of the HN output embeddings by comparing the performance when the suffix sequence is embedded with the original embeddings versus the HN-generated embeddings.

\section{Results and Discussion}
\label{sec:results}

\subsection{Encoder Models} 
Evaluation results for XLM-R with task adapters and  joint task and tokenization adapters, using different tokenization and embedding strategies are summarized in Table~\ref{table:xnli_task_adapter_original} for XNLI and Table~\ref{table:uner_task_adapter_original} for UNER. Figure~\ref{fig:subfig_xnli_13_lngs} illustrates the average performance across languages in XNLI with different sequence length reductions and adapters, while Figure~\ref{fig:subfig_xnli_eng_lng} focuses on English-only results. Corresponding results for UNER are shown in Figure~\ref{fig:UNER_f1_scores}. 

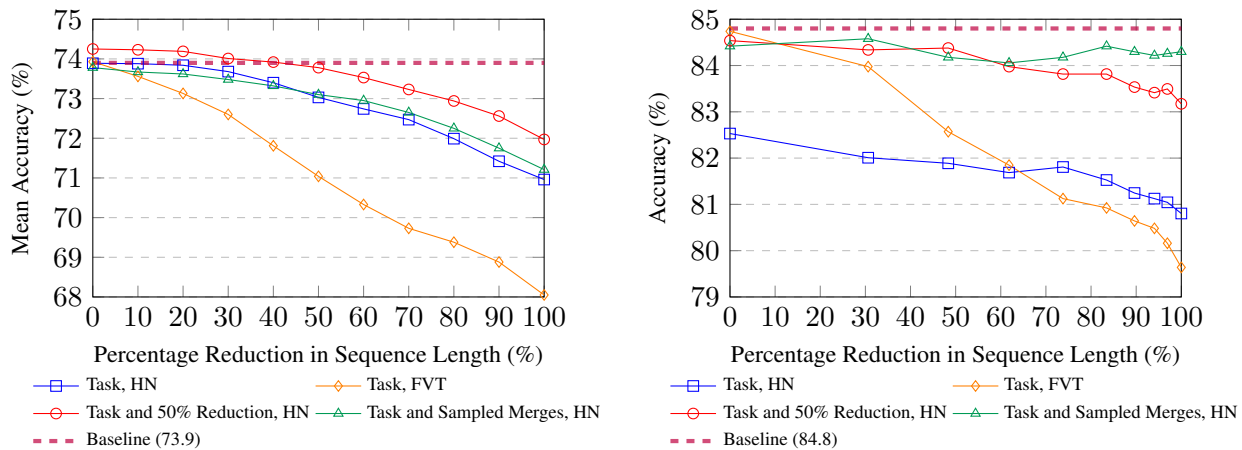
\begin{figure*}[t!]
\centering

\begin{subfigure}[b]{0.47\textwidth} 
\centering
\begin{tikzpicture}[scale=0.9]
\begin{axis}[
    width=1\columnwidth,
    height=0.7\columnwidth,
    xlabel={\small Percentage Reduction in Sequence Length (\%)},
    ylabel={\small Mean Accuracy (\%)},
    xmin=0, xmax=100,
    ymin=68, ymax=75,
    xtick={0,10,20,30,40,50,60,70,80,90,100},
    ytick={68,69,70,71,72,73,74,75},
    legend pos=south west,
    legend style={font=\scriptsize, cells={anchor=west}, draw=none, at={(0.5,-0.25)}, anchor=north, legend columns=2},
    ymajorgrids=true,
    grid style=dashed,
]
\addplot[
    color=blue,
    mark=square,
    ]
    coordinates {
    (0,73.89)(10,73.88)(20,73.84)(30,73.68)(40,73.40)(50,73.03)(60,72.74)(70,72.47)(80,71.99)(90,71.42)(100,70.96)
    };
    \addlegendentry{Task, HN}
\addplot[
    color=orange,
    mark=diamond,
    ]
    coordinates {
    (0,73.93)(10,73.56)(20,73.13)(30,72.60)(40,71.81)(50,71.04)(60,70.33)(70,69.73)(80,69.38)(90,68.88)(100,68.05)
    };
    \addlegendentry{Task, FVT}
\addplot[
    color=red,
    mark=o,
    ]
    coordinates {
    (0,74.25)(10,74.23)(20,74.19)(30,74.01)(40,73.92)(50,73.78)(60,73.53)(70,73.23)(80,72.94)(90,72.56)(100,71.97)
    };
    \addlegendentry{Task and 50\% Reduction, HN}
\addplot[
    color=ForestGreen,
    mark=triangle,
    ]
    coordinates {
    (0,73.78)(10,73.67)(20,73.62)(30,73.48)(40,73.32)(50,73.10)(60,72.95)(70,72.65)(80,72.25)(90,71.75)(100,71.21)
    };
    \addlegendentry{Task and Sampled Merges, HN}

\addplot[
    color=purple,
    ultra thick,
    dashed,
    opacity=0.7,
    domain=0:100,
] 
{73.9};
\addlegendentry{Baseline (73.9)}
\end{axis}
\end{tikzpicture}
\caption{Mean accuracies for XNLI across 13 languages.}
\label{fig:subfig_xnli_13_lngs}
\end{subfigure}
\hfill
\begin{subfigure}[b]{0.47\textwidth} 
\centering
\begin{tikzpicture}[scale=0.9]
\begin{axis}[
    width=\columnwidth,
    height=0.7\columnwidth,
    xlabel={\small Percentage Reduction in Sequence Length (\%)},
    ylabel={\small Accuracy (\%)},
    xmin=0, xmax=100,
    ymin=79, ymax=85,
    xtick={0, 10, 20, 30, 40, 50, 60, 70, 80, 90, 100},
    ytick={79,80,81,82,83,84,85},
    legend pos=south west,
    legend style={font=\scriptsize, cells={anchor=west}, draw=none, at={(0.5,-0.25)}, anchor=north, legend columns=2},
    ymajorgrids=true,
    grid style=dashed,
]
\addplot[
    color=blue,
    mark=square,
    ]
    coordinates {
    (0.000000,82.530120)(30.617923,82.008032)(48.359754,81.887550)(61.837593,81.686747)(73.788418,81.807229)
    (83.435781,81.526104)(89.627952,81.244980)(94.047234,81.124498)(96.913620,81.044177)(100.000000,80.803213)
    };
    \addlegendentry{Task, HN}
\addplot[
    color=orange,
    mark=diamond,
    ]
    coordinates {
    (0.000000,84.738956)(30.617923,83.975904)(48.359754,82.570281)(61.837593,81.847390)(73.788418,81.124498)
    (83.435781,80.923695)(89.627952,80.642570)(94.047234,80.481928)(96.913620,80.160643)(100.000000,79.638554)
    };
    \addlegendentry{Task, FVT}
\addplot[
    color=red,
    mark=o,
    ]
    coordinates {
    (0.000000,84.538153)(30.617923,84.337349)(48.359754,84.377510)(61.837593,83.975904)(73.788418,83.815261)
    (83.435781,83.815261)(89.627952,83.534137)(94.047234,83.413655)(96.913620,83.493976)(100.000000,83.172691)
    };
    \addlegendentry{Task and 50\% Reduction, HN}
\addplot[
    color=ForestGreen,
    mark=triangle,
    ]
    coordinates {
    (0.000000,84.417671)(30.617923,84.578313)(48.359754,84.176707)(61.837593,84.056225)(73.788418,84.176707)
    (83.435781,84.417671)(89.627952,84.297189)(94.047234,84.216867)(96.913620,84.257028)(100.000000,84.297189)
    };
    \addlegendentry{Task and Sampled Merges, HN}
\addplot[
    color=purple,
    ultra thick,
    dashed,
    opacity=0.7,
    domain=0:100,
] 
{84.8};
\addlegendentry{Baseline (84.8)}
\end{axis}
\end{tikzpicture}
\caption{Accuracies for different adapters on \textbf{English}.}
\label{fig:subfig_xnli_eng_lng}
\end{subfigure}
\vspace{-0.5mm}
\caption{Accuracies on \textbf{XNLI} with different adapters as a function of sequence length reduction (\%). Adapter names follow the format: \textit{Adapter type, Embeddings used}. 
{In this and subsequent figures, a 0\% reduction refers to sequence length obtained with the original subword-level tokenization, while 100\% indicates `upper bound' word-level tokenization. Intermediate percentages show proportional reductions between these two extremes.}}
\vspace{-1mm}
\end{figure*}
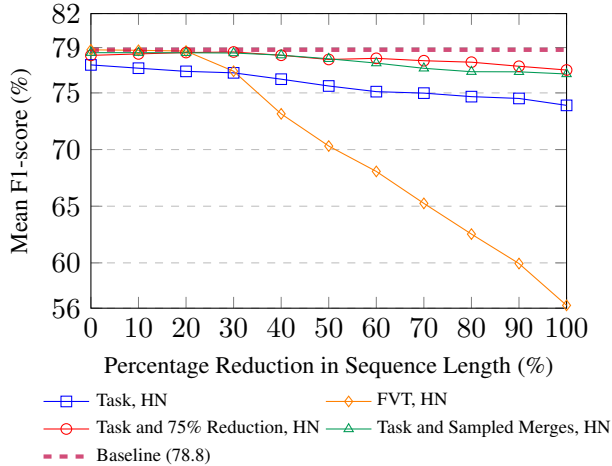
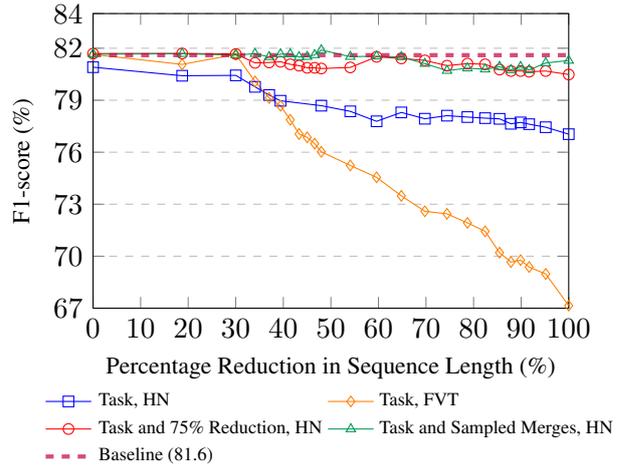
\begin{figure*}[!htbp]
\centering

\begin{subfigure}[b]{0.47\textwidth} 
\centering
\begin{tikzpicture}[scale=0.9]
\begin{axis}[
    width=\columnwidth, 
    height=0.7\columnwidth,
    xlabel={{\small Percentage Reduction in Sequence Length (\%)}},
    ylabel={\small Mean F1-score (\%)},
    xmin=0, xmax=100,
    ymin=56, ymax=82,
    xtick={0,10,20,30,40,50,60,70,80,90,100},
    ytick={56,60,65,70,75,79, 82},
    legend pos=south west,
    legend style={font=\scriptsize, cells={anchor=west}, draw=none, at={(0.5,-0.25)}, anchor=north, legend columns=2},
    ymajorgrids=true,
    grid style=dashed,
]
\addplot[
    color=blue,
    mark=square,
    ]
    coordinates {
    (0,77.47)(10,77.18)(20,76.90)(30,76.77)(40,76.21)(50,75.61)(60,75.11)(70,74.98)(80,74.67)(90,74.51)(100,73.90)
    };
    \addlegendentry{Task, HN}

\addplot[
    color=orange,
    mark=diamond,
    ]
    coordinates {
    (0,78.80)(10,78.76)(20,78.69)(30,76.91)(40,73.15)(50,70.31)(60,68.07)(70,65.25)(80,62.54)(90,59.94)(100,56.25)
    };
    \addlegendentry{Task, FVT}

\addplot[
    color=red,
    mark=o,
    ]
    coordinates {
    (0,78.31)(10,78.44)(20,78.56)(30,78.61)(40,78.32)(50,77.95)(60,78.05)(70,77.84)(80,77.71)(90,77.35)(100,77.02)
    };
    \addlegendentry{Task and 75\% Reduction, HN}

\addplot[
    color=ForestGreen,
    mark=triangle,
    ]
    coordinates {
    (0,78.55)(10,78.55)(20,78.54)(30,78.52)(40,78.33)(50,78.01)(60,77.63)(70,77.17)(80,76.87)(90,76.86)(100,76.67)
    };
    \addlegendentry{Task and Sampled Merges, HN}
    
\addplot[
    color=purple,
    ultra thick,
    dashed,
    opacity=0.7,
    domain=0:100,
] 
{78.8};
\addlegendentry{Baseline (78.8)}

\end{axis}
\end{tikzpicture}
\caption{Mean F1-scores for UNER across 5 languages.}
\end{subfigure}
\hfill
\begin{subfigure}[b]{0.47\textwidth} 
\centering
\begin{tikzpicture}[scale=0.9]
\begin{axis}[
    width=1.0\columnwidth, 
    height=0.7\columnwidth, 
    xlabel={\small Percentage Reduction in Sequence Length (\%)},
    ylabel={\small F1-score (\%)},
    xmin=0, xmax=100,
    ymin=67, ymax=84,
    xtick={0, 10, 20, 30, 40, 50, 60, 70, 80, 90, 100},
    ytick={67,70,73,76,79,82,84},
    legend pos=south west,
    legend style={font=\scriptsize, cells={anchor=west}, draw=none, at={(0.5,-0.25)}, anchor=north, legend columns=2}, 
    ymajorgrids=true,
    grid style=dashed,
]

\addplot[
    color=blue,
    mark=square,
    ]
    coordinates {
    (0.0, 80.905816)(18.737521, 80.412371)(30.010751, 80.4336602994321)(34.050069, 79.77296181630548)(37.029642, 79.29969104016478)
    (39.410229, 78.969072)(47.980341, 78.685157)(54.077715, 78.358974)(59.606819, 77.783479)(64.813393, 78.296562)
    (69.75887, 77.931388)(74.412533, 78.108941)(78.712947, 78.028747)(82.429734, 77.966102)(85.455383, 77.926078)
    (87.851329, 77.641026)(89.878667, 77.720739)(91.675626, 77.61807)(95.162033, 77.442455)(100.0, 77.040816)
    };
    \addlegendentry{Task, HN}

\addplot[
    color=orange,
    mark=diamond,
    ]
    coordinates {
    (0.0, 81.641026)(18.737521, 81.067214)(30.010751, 81.641026)(34.050069, 80.082559)(37.029642, 79.127726)
    (39.410229, 78.683386)(41.452926, 77.870564)(43.311319, 77.051751)(45.016127, 76.858639)(46.582706, 76.495278)
    (47.980341, 76.025237)(54.077715, 75.237092)(59.606819, 74.551214)(64.813393, 73.482428)(69.75887, 72.591006)
    (74.412533, 72.443488)(78.712947, 71.922246)(82.429734, 71.436314)(85.455383, 70.202296)(87.851329, 69.66046)
    (89.878667, 69.775096)(91.675626, 69.374314)(95.162033, 68.976898)(100.0, 67.146814)
    };
    \addlegendentry{Task, FVT}

\addplot[
    color=red,
    mark=o,
    ]
    coordinates {
    (0.0, 81.690141)(18.737521, 81.698686)(30.010751, 81.66582)(34.050069, 81.155015)(37.029642, 81.177067)
    (39.410229, 81.218274)(41.452926, 81.072874)(43.311319, 80.993411)(45.016127, 80.870445)(46.582706, 80.870445)
    (47.980341, 80.82954)(54.077715, 80.892043)(59.606819, 81.504065)(64.813393, 81.410569)(69.75887, 81.297516)
    (74.412533, 80.993411)(78.712947, 81.113924)(82.429734, 81.072874)(85.455383, 80.767289)(87.851329, 80.68583)
    (89.878667, 80.68583)(91.675626, 80.645161)(95.162033, 80.684105)(100.0, 80.482897)
    };
    \addlegendentry{Task and 75\% Reduction, HN}

\addplot[
    color=ForestGreen,
    mark=triangle,
    ]
    coordinates {
    (0.0, 81.657403)(18.737521, 81.684424)(30.010751, 81.605691)(34.050069, 81.684424)(37.029642, 81.500253)
    (39.410229, 81.661601)(41.452926, 81.657403)(43.311319, 81.496461)(45.016127, 81.515152)(46.582706, 81.612091)
    (47.980341, 81.896117)(54.077715, 81.515152)(59.606819, 81.537683)(64.813393, 81.496461)(69.75887, 81.111111)
    (74.412533, 80.707071)(78.712947, 80.848914)(82.429734, 80.788675)(85.455383, 80.930703)(87.851329, 80.745968)
    (89.878667, 80.928355)(91.675626, 80.745968)(95.162033, 81.149194)(100.0, 81.290973)
    };
    \addlegendentry{Task and Sampled Merges, HN}

\addplot[
    color=purple,
    ultra thick,
    dashed,
    opacity=0.7,
    domain=0:100,
] 
{81.6};
\addlegendentry{Baseline (81.6)}

\end{axis}
\end{tikzpicture}
\caption{F1-scores for different UNER adapters on \textbf{English}.}
\label{fig:subfig_uner_eng_lng}
\end{subfigure}
\caption{F1-scores on \textbf{UNER} with different adapters as a function of sequence length reduction (\%).}
\label{fig:UNER_f1_scores}
\vspace{-2mm}
\end{figure*}

\rparagraph{Dynamic tokenization substantially reduces sequence lengths} Applying dynamic tokenization with an adapter jointly trained for the task and a specified number of merges reduces token sequence length by an average of 22.5\% on the XNLI dataset (Table~\ref{table:xnli_task_adapter_original}), with an average accuracy decrease of 1.9\% compared to the original tokenization and embeddings. On the UNER dataset (Table~\ref{table:uner_task_adapter_original}), this approach achieves a 26.4\% reduction in sequence length, with only a 1.7\% decrease in F1-score.

\rparagraph{Sampling the tokenization granularity improves in-domain performance} Comparing our two types of adapters, we find that, for both datasets, the adapter trained with $m$ sampled from $\mathrm{U}(0, m_{\text{max}})$ on average outperforms the fixed-merge adapter on English (i.e., in-domain). This adapter yields better results across all sequence length reductions and nearly closes the gap with the baseline performance with the original tokenization and embeddings. For XNLI, with word-level tokenization, it obtains an accuracy of 84.3\%, compared to 83.2\% with the fixed-merge adapter and 84.8\% with the baseline (Figure~\ref{fig:subfig_xnli_eng_lng}). These results align with the UNER results where the adapter trained with sampled merges achieves an F1-score of 81.3\% on word-level tokenization, compared with the fixed-merge adapter at 80.5\% and close to the baseline of 81.6\% (Figure~\ref{fig:subfig_uner_eng_lng}), confirming that the model benefits from a balanced exposure to different tokenization granularities on in-domain tasks.

{\rparagraph{Cross-lingual performance} Unlike our in-domain results, the fixed-merges adapter consistently shows stronger cross-lingual transferability than its counterpart trained with sampled merges. On XNLI, it obtains an average accuracy of 72.0\% compared to 71.2\% 
(Table~\ref{table:xnli_task_adapter_original}). Similarly, for UNER, it achieves 77.0\% compared to 76.7\% (Table~\ref{table:uner_task_adapter_original}). 


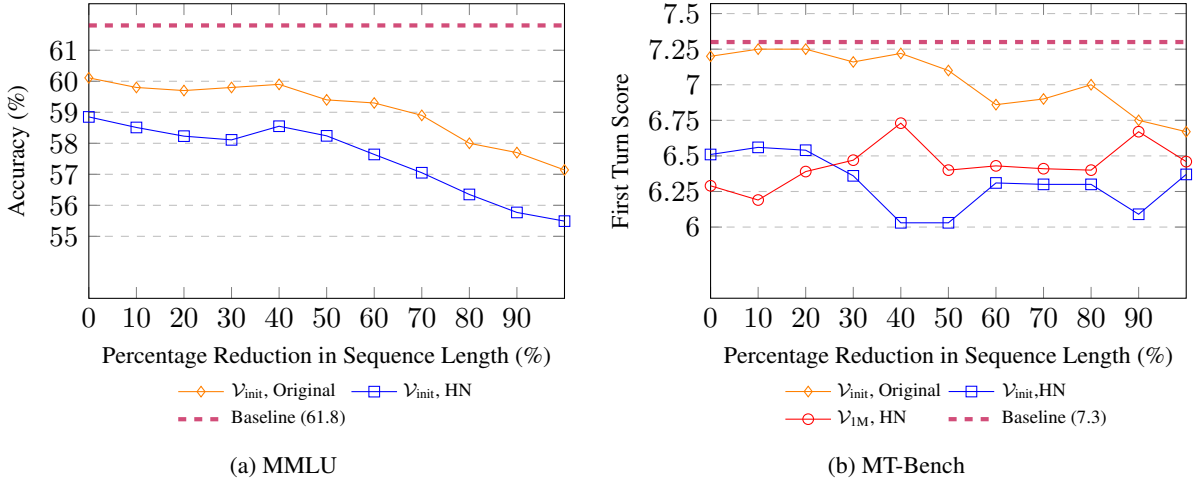
\begin{figure*}[!t]
\centering




\begin{subfigure}[b]{0.47\textwidth}
\centering
\begin{tikzpicture}[scale=0.9]
\begin{axis}[
    width=\columnwidth,
    height=0.7\columnwidth,
    xlabel={\small Percentage Reduction in Sequence Length (\%)},
    ylabel={\small Accuracy (\%)},
    xmin=0, xmax=100,
    ymin=35, ymax=63,               
    xtick={0,10,20,30,40,50,60,70,80,90,100},
    ytick={35,40,45,50,55,60},
    legend pos=south west,
    legend style={
        font=\scriptsize,
        cells={anchor=west},
        draw=none,
        at={(0.5,-0.25)},
        anchor=north,
        legend columns=2
    },
    ymajorgrids=true,
    grid style=dashed,
]
\addplot[
    color=orange,
    mark=diamond,
]
coordinates {
    (0, 61.22)
    (10, 57.76)
    (20, 56.36)
    (30, 54.62)
    (40, 53.32)
    (50, 52.24)
    (60, 50.36)
    (70, 43.78)
    (80, 38.12)
    (90, 38.08)
    (100, 37.14)
};
\addlegendentry{FVT, Original}

\addplot[
    color=blue,
    mark=square,
]
coordinates {
    (0, 57.34)
    (10, 56.82)
    (20, 55.88)
    (30, 56.02)
    (40, 55.56)
    (50, 55.28)
    (60, 53.20)
    (70, 51.46)
    (80, 47.38)
    (90, 48.12)
    (100, 48.22)
};
\addlegendentry{HN, Original}

\addplot[
    color=ForestGreen,
    mark=triangle,
]
coordinates {
    (0, 56.34)
    (10, 55.58)
    (20, 55.02)
    (30, 54.38)
    (40, 54.22)
    (50, 55.00)
    (60, 51.96)
    (70, 48.56)
    (80, 46.36)
    (90, 45.88)
    (100, 46.22)
};
\addlegendentry{HN, HN}

\addplot[
    color=purple,
    ultra thick,
    dashed,
    opacity=0.7,
    domain=0:100,
]
{61.2};
\addlegendentry{Baseline (61.2)}

\end{axis}
\end{tikzpicture}
\caption{MMLU, average accuracies across 5 languages. Average sequence length at 0\%: 905.9 and 100\%: 598.1}
\label{fig:prefilling_mmlu}
\end{subfigure}
\hfill
\begin{subfigure}[b]{0.47\textwidth} 
\centering
\begin{tikzpicture}[scale=0.9]
\begin{axis}[
    width=\columnwidth, 
    height=0.7\columnwidth,
    xlabel={\small Percentage Reduction in Sequence Length (\%)},
    ylabel={\small First Turn Score},
    xmin=0, xmax=100,
    ymin=5.5, ymax=7.57,
    xtick={0,10,20,30,40,50,60,70,80,90},
    ytick={6.0,6.25,6.5,6.75,7.0,7.25,7.5},
    legend pos=south west,
    legend style={font=\scriptsize, cells={anchor=west}, draw=none, at={(0.5,-0.25)}, anchor=north, legend columns=2},
    ymajorgrids=true,
    grid style=dashed,
]

\addplot[
    color=blue,
    mark=diamond,
    ]
    coordinates {
    (0,7.20)
    (10,7.25)
    (20,7.25)
    (30,7.16)
    (40,7.22)
    (50,7.10)
    (60,6.86)
    (70,6.90)
    (80,7.00)
    (90,6.75)
    (100,6.67)
    };
    \addlegendentry{$\vcal_\text{init}$, Original}
    
\addplot[
    color=ForestGreen,
    mark=triangle,
    ]
    coordinates {
    (0,6.51)
    (10,6.56)
    (20,6.54)
    (30,6.36)
    (40,6.03)
    (50,6.03)
    (60,6.31)
    (70,6.30)
    (80,6.30)
    (90,6.09)
    (100,6.37)
    };
    \addlegendentry{$\vcal_\text{init}$,HN}

\addplot[
    color=red,
    mark=o,
    ]
    coordinates {
    (0,6.29)
    (10,6.19)
    (20,6.39)
    (30,6.47)
    (40,6.73)
    (50,6.40)
    (60,6.43)
    (70,6.41)
    (80,6.40)
    (90,6.67)
    (100,6.46)
    };
    \addlegendentry{$\vcalmil$, HN }

\addplot[
    color=purple,
    ultra thick,
    dashed,
    opacity=0.7,
    domain=0:100,
] 
{7.3};
\addlegendentry{Baseline (7.3)}

\end{axis}
\end{tikzpicture}
\caption{MT-Bench, English. Average sequence length at 0\%: 804.4 and 100\%: 692.0}
\label{fig:prefilling_mt_bench}
\end{subfigure}
\vspace{-1mm}
\caption{Performance of dynamic tokenization applied to decoders for scoring and prefilling, evaluated across various vocabularies and embeddings in a zero-shot setting. For MMLU, the \textit{baseline} is the accuracy with the original tokenization and embeddings, while for MT-Bench, it is the first-turn score with the original setup. Labels follow the format: a) MMLU: Embeddings used for the dynamic tokenized input and output embeddings used for scoring; b) MT-Bench: Vocabulary $\vcal_\text{gen}$ and embeddings used for generation. $\vcalmil$ expands $\vcalinit$ to 1M tokens.}
\label{fig:decoders_prefilling}
\vspace{-1.5mm}
\end{figure*}

\rparagraph{Heuristic embeddings $\ll$ HN-generated embeddings $<$ original embeddings} Using original tokenization with HN embeddings (Setting 2 in Table~\ref{table:xnli_task_adapter_original}) shows comparable results to original embeddings (Setting 1), in both English and cross-lingual contexts, highlighting the quality of HN embeddings. However, there is a noticeable gap between subword- and word-level HN embeddings in Settings 1 and 3, as the model was not previously exposed to HN embeddings. FVT embeddings, by contrast, show a prominent performance drop: for instance, FVT achieves an average accuracy of 68.2\% on XNLI, compared to 71.1\% (Table~\ref{table:xnli_task_adapter_original}) with word-level HN embeddings, and scores 56.3\% F1 on UNER, substantially lower than the 74.6\% achieved with HN embeddings (Table~\ref{table:uner_task_adapter_original}). This suggests that HN embeddings more effectively capture the semantic nuances required for tasks like NER, where accurate token representation is important.

\subsection{Decoder Models}\label{sec:results_decoders}
Figure~\ref{fig:decoders_prefilling} presents performance trends for different granularities obtained using dynamic tokenization for scoring and prefilling (see \Cref{sec:decoder_experiments}). 

\rparagraph{Dynamic tokenization improves prefilling and scoring efficiency} In zero-shot evaluation of dynamic tokenization across different granularities, we observe that using the original \textit{output embeddings} consistently outperform the HN-generated output embeddings, highlighting that the degradation in performance is primarily due to the type of output embeddings used rather than the input embeddings (Figure~\ref{fig:decoders_prefilling}). 
For both MMLU and MT-Bench, reducing the sequence length by 40-50\% (relative to the word-level) --- corresponding to a 17\% average reduction for MMLU and $\approx6\%$ for MT-Bench (in absolute terms) --- and using the original output embeddings (from $\vcalinit$) yields the smallest gap to the baseline, unlike FVT, whose performance quickly degrades above 30\% reduction. Overall, we can use this insight to compress the key-value cache with minimal performance degradation by exclusively changing the input embeddings i.e., without \textit{any} changes to the pre-trained model.

\rparagraph{Expanding the vocabulary allows achieving some of the benefits of dynamic tokenization for autoregressive generation} Our dynamic tokenization approach works for LM \textit{scoring} (i.e., computing the conditional probability of a text) and \textit{prefilling}, as we know the sequence in advance. However, this is not the case for autoregressive generation. To address this, we propose a method that expands the initial vocabulary, $\vcalinit$, from $\approx$32k tokens to 1M tokens, $\vcalmil$, moving token granularity closer to word-level while maintaining a large bounded vocabulary. In a nutshell, our approach involves three steps: \emph{(1)} expanding $\vcalinit$ to $\vcalmil$ by applying BPE on a large corpus; \emph{(2)} using longest-prefix (LP) tokenization --- instead of the previous dynamic tokenization --- to overcome the challenges involved in merging two tokenizers (i.e., the initial tokenizer and the one obtained for $\vcalmil$); and \emph{(3)} constructing an approximate nearest neighbor index to efficiently retrieve token embeddings using an HN.
This approach is described in full technical detail in Appendix~\ref{sec:vocabulary_expasion_decoders} and Tables~\ref{table:mmlu_results} and~\ref{table:mtbench_results} summarize the results with different settings. 

Using this expanded vocabulary we achieve shorter token sequences --- similar to word-level tokenization --- at the expense of degrading the performance by 5.9\% for MMLU (English) and 0.9 points on MT-Bench (Table~\ref{table:mmlu_results} and Table~\ref{table:mtbench_results}). Additionally, retrieving the embeddings for tokens in $\vcalmil$ on-the-fly using an HN allows us to maintain the original model's parameter count. The current performance gap could potentially be minimized through n-shot tokenizer transfer, as demonstrated by~\citet{minixhofer2024zero}, but we leave this exploration beyond zero-shot setups to future research. 


\begin{table}[!htbp]
\centering
\scriptsize
\begin{tabular}{l l c c c}
\toprule
\multirow{2}{*}{\textbf{Lng}} & 
\multirow{2}{*}{\textbf{FLOPs}} & 
\multicolumn{3}{c}{\textbf{Sequence Reduction}} \\
\cmidrule(lr){3-5}
 & & \textbf{0\%} & \textbf{50\%} & \textbf{100\%} \\
\midrule
\multirow{2}{*}{\textbf{en}}

   & Model    & 10.1T & 9.4T & 8.5T \\
             & Hypernet & 169.3B (1.7\%) & 191.0B (2.0\%) & 199.8B (2.2\%) \\
\midrule
\multirow{2}{*}{\textbf{fr}}

   & Model    & 14.4T & 12.2T & 9.7T \\
            & Hypernet & 91.8B (0.6\%) & 163.5B (1.3\%) & 238.5B (2.4\%) \\
\bottomrule
\end{tabular}
\caption{\textit{FLOPs per sample} with dynamic tokenization on multilingual MMLU. Percentages in parentheses denote the fraction of hypernetwork FLOPs out of total.}
\label{tab:flops_comparison_main}
\end{table}

\rparagraph{Throughput analysis} Table \ref{tab:flops_comparison_main} shows that dynamic tokenization reduces the main model’s FLOPs (e.g., 14.4T to 9.7T in French, 67.4\% of the baseline) while the hypernetwork’s FLOPs remain below 3\% of the total.
The gains align with sequence reduction, yielding near-linear throughput improvements with negligible overhead from the hypernetwork. Appendix~\ref{sec:throughput_analysis} shows the complete set of results.





\section{Conclusion}
We proposed a novel dynamic tokenization method for (large) language models, using a hypernetwork to dynamically adapt token boundaries based on the input data, which efficiently generates token embeddings on-the-fly. We then demonstrated its usefulness both on encoder- and decoder-style models. As some main findings, we highlight that for encoder models (e.g., XLM-R), our approach substantially reduces token sequence lengths by $>20\%$ on average over 14 languages, with less than 2\% loss in accuracy. When applied to decoder-style models (e.g., Mistral-7B) for prefilling and scoring, our method yields minimal performance degradation with up to 6\% reduction in (absolute) sequence length on English. Overall, these results demonstrate that dynamic tokenization can mitigate some of the limitations of static tokenizers, particularly in multilingual settings, improving inference efficiency and promoting fairness across languages. 

\section*{Limitations}

There is computational overhead associated with \emph{(1)} generating a new vocabulary for each batch and \emph{(2)} on-the-fly generation of token embeddings using an HN \citep[c.f.,][]{minixhofer2024zero}. To amortize the latter, we implemented an HN embeddings cache, particularly motivated by the frequent repetition of certain tokens (e.g., ``the''; c.f., Appendix~\ref{sec:emb_cache}). Furthermore, the success of on-the-fly token embeddings relies on the accuracy and robustness of the hypernetwork, and requires a pretrained hypernetwork (c.f. Appendix~\ref{sec:hypernetwork_training}). Finally, our current dynamic tokenization approach is primarily designed for settings where the full sequence is known in advance (e.g., scoring and prefilling in decoder models). For autoregressive generation, we offered a first solution which relies on a large but static, bounded vocabulary to achieve some of the benefits of dynamic tokenization (e.g., token sequence length compression). Closing this gap by integrating `true' dynamic tokenization into autoregressive generation remains an open challenge for future research.


\section*{Acknowledgments}

This work has been supported by a Royal Society University Research Fellowship \textit{‘Inclusive and Sustainable Language Technology for a Truly Multilingual World’} (no 221137; 2022-) awarded to Ivan Vuli\'{c}. We thank Google for sponsoring our attendance at ACL 2025.

\bibliography{anthology,cites}

\begin{thebibliography}{49}
\providecommand{\natexlab}[1]{#1}

\bibitem[{Ahia et~al.(2023)Ahia, Kumar, Gonen, Kasai, Mortensen, Smith, and Tsvetkov}]{ahia2023all}
Orevaoghene Ahia, Sachin Kumar, Hila Gonen, Jungo Kasai, David Mortensen, Noah Smith, and Yulia Tsvetkov. 2023.
\newblock \href {https://doi.org/10.18653/v1/2023.emnlp-main.614} {Do all languages cost the same? tokenization in the era of commercial language models}.
\newblock In \emph{Proceedings of the 2023 Conference on Empirical Methods in Natural Language Processing}, pages 9904--9923, Singapore. Association for Computational Linguistics.

\bibitem[{Ali et~al.(2024)Ali, Fromm, Thellmann, Rutmann, L{\"u}bbering, Leveling, Klug, Ebert, Doll, Buschhoff, Jain, Weber, Jurkschat, Abdelwahab, John, Ortiz~Suarez, Ostendorff, Weinbach, Sifa, Kesselheim, and Flores-Herr}]{ali2023tokenizer}
Mehdi Ali, Michael Fromm, Klaudia Thellmann, Richard Rutmann, Max L{\"u}bbering, Johannes Leveling, Katrin Klug, Jan Ebert, Niclas Doll, Jasper Buschhoff, Charvi Jain, Alexander Weber, Lena Jurkschat, Hammam Abdelwahab, Chelsea John, Pedro Ortiz~Suarez, Malte Ostendorff, Samuel Weinbach, Rafet Sifa, Stefan Kesselheim, and Nicolas Flores-Herr. 2024.
\newblock \href {https://aclanthology.org/2024.findings-naacl.247} {Tokenizer choice for {LLM} training: Negligible or crucial?}
\newblock In \emph{Findings of the Association for Computational Linguistics: NAACL 2024}, pages 3907--3924, Mexico City, Mexico. Association for Computational Linguistics.

\bibitem[{Bostrom and Durrett(2020)}]{bostrom2020byte}
Kaj Bostrom and Greg Durrett. 2020.
\newblock \href {https://doi.org/10.18653/v1/2020.findings-emnlp.414} {Byte pair encoding is suboptimal for language model pretraining}.
\newblock In \emph{Findings of the Association for Computational Linguistics: EMNLP 2020}, pages 4617--4624, Online. Association for Computational Linguistics.

\bibitem[{Chiang and Lee(2023)}]{chiang2023can}
Cheng-Han Chiang and Hung-yi Lee. 2023.
\newblock \href {https://doi.org/10.18653/v1/2023.acl-long.870} {Can large language models be an alternative to human evaluations?}
\newblock In \emph{Proceedings of the 61st Annual Meeting of the Association for Computational Linguistics (Volume 1: Long Papers)}, pages 15607--15631, Toronto, Canada. Association for Computational Linguistics.

\bibitem[{Clark et~al.(2022)Clark, Garrette, Turc, and Wieting}]{clark2022canine}
Jonathan~H. Clark, Dan Garrette, Iulia Turc, and John Wieting. 2022.
\newblock \href {https://doi.org/10.1162/tacl_a_00448} {Canine: Pre-training an efficient tokenization-free encoder for language representation}.
\newblock \emph{Transactions of the Association for Computational Linguistics}, 10:73--91.

\bibitem[{Conneau et~al.(2020)Conneau, Khandelwal, Goyal, Chaudhary, Wenzek, Guzm{\'a}n, Grave, Ott, Zettlemoyer, and Stoyanov}]{conneau2020unsupervised}
Alexis Conneau, Kartikay Khandelwal, Naman Goyal, Vishrav Chaudhary, Guillaume Wenzek, Francisco Guzm{\'a}n, Edouard Grave, Myle Ott, Luke Zettlemoyer, and Veselin Stoyanov. 2020.
\newblock \href {https://doi.org/10.18653/v1/2020.acl-main.747} {Unsupervised cross-lingual representation learning at scale}.
\newblock In \emph{Proceedings of the 58th Annual Meeting of the Association for Computational Linguistics}, pages 8440--8451, Online. Association for Computational Linguistics.

\bibitem[{Conneau et~al.(2018)Conneau, Rinott, Lample, Williams, Bowman, Schwenk, and Stoyanov}]{conneau2018xnli}
Alexis Conneau, Ruty Rinott, Guillaume Lample, Adina Williams, Samuel Bowman, Holger Schwenk, and Veselin Stoyanov. 2018.
\newblock \href {https://doi.org/10.18653/v1/D18-1269} {{XNLI}: Evaluating cross-lingual sentence representations}.
\newblock In \emph{Proceedings of the 2018 Conference on Empirical Methods in Natural Language Processing}, pages 2475--2485, Brussels, Belgium. Association for Computational Linguistics.

\bibitem[{Dagan et~al.(2024)Dagan, Synnaeve, and Rozi{\`e}re}]{dagan2024getting}
Gautier Dagan, Gabriel Synnaeve, and Baptiste Rozi{\`e}re. 2024.
\newblock \href {https://arxiv.org/abs/2402.01035} {Getting the most out of your tokenizer for pre-training and domain adaptation}.
\newblock \emph{ArXiv preprint}, abs/2402.01035.

\bibitem[{Devlin et~al.(2019)Devlin, Chang, Lee, and Toutanova}]{devlin2018bert}
Jacob Devlin, Ming-Wei Chang, Kenton Lee, and Kristina Toutanova. 2019.
\newblock \href {https://doi.org/10.18653/v1/N19-1423} {{BERT}: Pre-training of deep bidirectional transformers for language understanding}.
\newblock In \emph{Proceedings of the 2019 Conference of the North {A}merican Chapter of the Association for Computational Linguistics: Human Language Technologies, Volume 1 (Long and Short Papers)}, pages 4171--4186, Minneapolis, Minnesota. Association for Computational Linguistics.

\bibitem[{Dobler and de~Melo(2023)}]{dobler2023focus}
Konstantin Dobler and Gerard de~Melo. 2023.
\newblock \href {https://doi.org/10.18653/v1/2023.emnlp-main.829} {{FOCUS}: Effective embedding initialization for monolingual specialization of multilingual models}.
\newblock In \emph{Proceedings of the 2023 Conference on Empirical Methods in Natural Language Processing}, pages 13440--13454, Singapore. Association for Computational Linguistics.

\bibitem[{El~Boukkouri et~al.(2020)El~Boukkouri, Ferret, Lavergne, Noji, Zweigenbaum, and Tsujii}]{boukkouri2020characterbert}
Hicham El~Boukkouri, Olivier Ferret, Thomas Lavergne, Hiroshi Noji, Pierre Zweigenbaum, and Jun{'}ichi Tsujii. 2020.
\newblock \href {https://doi.org/10.18653/v1/2020.coling-main.609} {{C}haracter{BERT}: Reconciling {ELM}o and {BERT} for word-level open-vocabulary representations from characters}.
\newblock In \emph{Proceedings of the 28th International Conference on Computational Linguistics}, pages 6903--6915, Barcelona, Spain (Online). International Committee on Computational Linguistics.

\bibitem[{Fujii et~al.(2023)Fujii, Shibata, Yamaguchi, Morishita, and Sogawa}]{fujii2023different}
Takuro Fujii, Koki Shibata, Atsuki Yamaguchi, Terufumi Morishita, and Yasuhiro Sogawa. 2023.
\newblock \href {https://doi.org/10.18653/v1/2023.acl-srw.5} {How do different tokenizers perform on downstream tasks in scriptio continua languages?: A case study in {J}apanese}.
\newblock In \emph{Proceedings of the 61st Annual Meeting of the Association for Computational Linguistics (Volume 4: Student Research Workshop)}, pages 39--49, Toronto, Canada. Association for Computational Linguistics.

\bibitem[{Gee et~al.(2022)Gee, Zugarini, Rigutini, and Torroni}]{gee2024fast}
Leonidas Gee, Andrea Zugarini, Leonardo Rigutini, and Paolo Torroni. 2022.
\newblock \href {https://doi.org/10.18653/v1/2022.emnlp-industry.41} {Fast vocabulary transfer for language model compression}.
\newblock In \emph{Proceedings of the 2022 Conference on Empirical Methods in Natural Language Processing: Industry Track}, pages 409--416, Abu Dhabi, UAE. Association for Computational Linguistics.

\bibitem[{Golkar et~al.(2023)Golkar, Pettee, Eickenberg, Bietti, Cranmer, Krawezik, Lanusse, McCabe, Ohana, Parker, Régaldo-Saint~Blancard, Tesileanu, Cho, and Ho}]{golkar2023xval}
Siavash Golkar, Mariel Pettee, Michael Eickenberg, Alberto Bietti, Miles Cranmer, Geraud Krawezik, Francois Lanusse, Michael McCabe, Ruben Ohana, Liam Parker, Bruno Régaldo-Saint~Blancard, Tiberiu Tesileanu, Kyunghyun Cho, and Shirley Ho. 2023.
\newblock \href {https://arxiv.org/abs/2310.02989} {{xVal: A continuous number encoding for large language models}}.
\newblock \emph{ArXiv preprint}, abs/2310.02989.

\bibitem[{Guo et~al.(2020)Guo, Sun, Lindgren, Geng, Simcha, Chern, and Kumar}]{guo2020accelerating}
Ruiqi Guo, Philip Sun, Erik Lindgren, Quan Geng, David Simcha, Felix Chern, and Sanjiv Kumar. 2020.
\newblock \href {http://proceedings.mlr.press/v119/guo20h.html} {Accelerating large-scale inference with anisotropic vector quantization}.
\newblock In \emph{Proceedings of the 37th International Conference on Machine Learning, {ICML} 2020, 13-18 July 2020, Virtual Event}, volume 119 of \emph{Proceedings of Machine Learning Research}, pages 3887--3896. {PMLR}.

\bibitem[{Hofmann et~al.(2022)Hofmann, Schuetze, and Pierrehumbert}]{hofmann2022embarrassingly}
Valentin Hofmann, Hinrich Schuetze, and Janet Pierrehumbert. 2022.
\newblock \href {https://doi.org/10.18653/v1/2022.acl-short.43} {An embarrassingly simple method to mitigate undesirable properties of pretrained language model tokenizers}.
\newblock In \emph{Proceedings of the 60th Annual Meeting of the Association for Computational Linguistics (Volume 2: Short Papers)}, pages 385--393, Dublin, Ireland. Association for Computational Linguistics.

\bibitem[{Hu et~al.(2022)Hu, Shen, Wallis, Allen{-}Zhu, Li, Wang, Wang, and Chen}]{hu2021lora}
Edward~J. Hu, Yelong Shen, Phillip Wallis, Zeyuan Allen{-}Zhu, Yuanzhi Li, Shean Wang, Lu~Wang, and Weizhu Chen. 2022.
\newblock \href {https://openreview.net/forum?id=nZeVKeeFYf9} {Lora: Low-rank adaptation of large language models}.
\newblock In \emph{The Tenth International Conference on Learning Representations, {ICLR} 2022, Virtual Event, April 25-29, 2022}. OpenReview.net.

\bibitem[{Jiang et~al.(2023)Jiang, Sablayrolles, Mensch, Bamford, Chaplot, Casas, Bressand, Lengyel, Lample, Saulnier et~al.}]{jiang2023mistral}
Albert~Q Jiang, Alexandre Sablayrolles, Arthur Mensch, Chris Bamford, Devendra~Singh Chaplot, Diego de~las Casas, Florian Bressand, Gianna Lengyel, Guillaume Lample, Lucile Saulnier, et~al. 2023.
\newblock \href {https://arxiv.org/abs/2310.06825} {Mistral 7b}.
\newblock \emph{ArXiv preprint}, abs/2310.06825.

\bibitem[{Kudo(2018)}]{kudo2018subword}
Taku Kudo. 2018.
\newblock \href {https://doi.org/10.18653/v1/P18-1007} {Subword regularization: Improving neural network translation models with multiple subword candidates}.
\newblock In \emph{Proceedings of the 56th Annual Meeting of the Association for Computational Linguistics (Volume 1: Long Papers)}, pages 66--75.

\bibitem[{Kudo and Richardson(2018)}]{kudo2018sentencepiece}
Taku Kudo and John Richardson. 2018.
\newblock \href {https://doi.org/10.18653/v1/D18-2012} {{S}entence{P}iece: A simple and language independent subword tokenizer and detokenizer for neural text processing}.
\newblock In \emph{Proceedings of the 2018 Conference on Empirical Methods in Natural Language Processing: System Demonstrations}, pages 66--71, Brussels, Belgium. Association for Computational Linguistics.

\bibitem[{Kudugunta et~al.(2023)Kudugunta, Caswell, Zhang, Garcia, Xin, Kusupati, Stella, Bapna, and Firat}]{kudugunta2024madlad}
Sneha Kudugunta, Isaac Caswell, Biao Zhang, Xavier Garcia, Derrick Xin, Aditya Kusupati, Romi Stella, Ankur Bapna, and Orhan Firat. 2023.
\newblock \href {http://papers.nips.cc/paper\_files/paper/2023/hash/d49042a5d49818711c401d34172f9900-Abstract-Datasets\_and\_Benchmarks.html} {{MADLAD-400:} {A} multilingual and document-level large audited dataset}.
\newblock In \emph{Advances in Neural Information Processing Systems 36: Annual Conference on Neural Information Processing Systems 2023, NeurIPS 2023, New Orleans, LA, USA, December 10 - 16, 2023}.

\bibitem[{Lan et~al.(2023)Lan, Cai, Wang, Huang, and Mao}]{lan2023copyneed}
Tian Lan, Deng Cai, Yan Wang, Heyan Huang, and Xian{-}Ling Mao. 2023.
\newblock \href {https://openreview.net/pdf?id=CROlOA9Nd8C} {Copy is all you need}.
\newblock In \emph{The Eleventh International Conference on Learning Representations, {ICLR} 2023, Kigali, Rwanda, May 1-5, 2023}. OpenReview.net.

\bibitem[{Li et~al.(2024)Li, Chen, Holtzman, Chen, Lin, Yih, and Lin}]{li2024nearest}
Minghan Li, Xilun Chen, Ari Holtzman, Beidi Chen, Jimmy Lin, Wen-tau Yih, and Xi~Victoria Lin. 2024.
\newblock \href {https://arxiv.org/abs/2405.19325} {Nearest neighbor speculative decoding for llm generation and attribution}.
\newblock \emph{ArXiv preprint}, abs/2405.19325.

\bibitem[{Liang et~al.(2023)Liang, Gonen, Mao, Hou, Goyal, Ghazvininejad, Zettlemoyer, and Khabsa}]{liang2023xlm}
Davis Liang, Hila Gonen, Yuning Mao, Rui Hou, Naman Goyal, Marjan Ghazvininejad, Luke Zettlemoyer, and Madian Khabsa. 2023.
\newblock \href {https://doi.org/10.18653/v1/2023.emnlp-main.813} {{XLM}-{V}: Overcoming the vocabulary bottleneck in multilingual masked language models}.
\newblock In \emph{Proceedings of the 2023 Conference on Empirical Methods in Natural Language Processing}, pages 13142--13152, Singapore. Association for Computational Linguistics.

\bibitem[{Limisiewicz et~al.(2024)Limisiewicz, Blevins, Gonen, Ahia, and Zettlemoyer}]{limisiewicz-etal-2024-myte}
Tomasz Limisiewicz, Terra Blevins, Hila Gonen, Orevaoghene Ahia, and Luke Zettlemoyer. 2024.
\newblock \href {https://doi.org/10.18653/v1/2024.acl-long.804} {{MYTE}: Morphology-driven byte encoding for better and fairer multilingual language modeling}.
\newblock In \emph{Proceedings of the 62nd Annual Meeting of the Association for Computational Linguistics (Volume 1: Long Papers)}, pages 15059--15076, Bangkok, Thailand. Association for Computational Linguistics.

\bibitem[{Liu et~al.(2024)Liu, Lin, Wang, and Schuetze}]{liu2024ofa}
Yihong Liu, Peiqin Lin, Mingyang Wang, and Hinrich Schuetze. 2024.
\newblock \href {https://aclanthology.org/2024.findings-naacl.68} {{OFA}: A framework of initializing unseen subword embeddings for efficient large-scale multilingual continued pretraining}.
\newblock In \emph{Findings of the Association for Computational Linguistics: NAACL 2024}, pages 1067--1097, Mexico City, Mexico. Association for Computational Linguistics.

\bibitem[{Mayhew et~al.(2024)Mayhew, Blevins, Liu, Suppa, Gonen, Imperial, Karlsson, Lin, Ljube{\v{s}}i{\'c}, Miranda, Plank, Riabi, and Pinter}]{mayhew2023universal}
Stephen Mayhew, Terra Blevins, Shuheng Liu, Marek Suppa, Hila Gonen, Joseph~Marvin Imperial, B{\"o}rje Karlsson, Peiqin Lin, Nikola Ljube{\v{s}}i{\'c}, Lester~James Miranda, Barbara Plank, Arij Riabi, and Yuval Pinter. 2024.
\newblock \href {https://aclanthology.org/2024.naacl-long.243} {Universal {NER}: A gold-standard multilingual named entity recognition benchmark}.
\newblock In \emph{Proceedings of the 2024 Conference of the North American Chapter of the Association for Computational Linguistics: Human Language Technologies (Volume 1: Long Papers)}, pages 4322--4337, Mexico City, Mexico. Association for Computational Linguistics.

\bibitem[{Mielke et~al.(2021)Mielke, Alyafeai, Salesky, Raffel, Dey, Gall{\'e}, Raja, Si, Lee, Sagot et~al.}]{mielke2021between}
Sabrina~J Mielke, Zaid Alyafeai, Elizabeth Salesky, Colin Raffel, Manan Dey, Matthias Gall{\'e}, Arun Raja, Chenglei Si, Wilson~Y Lee, Beno\^{i}t Sagot, et~al. 2021.
\newblock \href {https://arxiv.org/abs/2112.10508} {{Between words and characters: A brief history of open-vocabulary modeling and tokenization in NLP}}.
\newblock \emph{ArXiv preprint}, abs/2112.10508.

\bibitem[{Minaee et~al.(2024)Minaee, Mikolov, Nikzad, Chenaghlu, Socher, Amatriain, and Gao}]{minaee2024large}
Shervin Minaee, Tomas Mikolov, Narjes Nikzad, Meysam Chenaghlu, Richard Socher, Xavier Amatriain, and Jianfeng Gao. 2024.
\newblock \href {https://arxiv.org/abs/2402.06196} {{Large Language Models: A Survey}}.
\newblock \emph{ArXiv preprint}, abs/2402.06196.

\bibitem[{Minixhofer et~al.(2022)Minixhofer, Paischer, and Rekabsaz}]{minixhofer2022wechsel}
Benjamin Minixhofer, Fabian Paischer, and Navid Rekabsaz. 2022.
\newblock \href {https://doi.org/10.18653/v1/2022.naacl-main.293} {{WECHSEL}: Effective initialization of subword embeddings for cross-lingual transfer of monolingual language models}.
\newblock In \emph{Proceedings of the 2022 Conference of the North American Chapter of the Association for Computational Linguistics: Human Language Technologies}, pages 3992--4006, Seattle, United States. Association for Computational Linguistics.

\bibitem[{Minixhofer et~al.(2024)Minixhofer, Ponti, and Vuli{\'c}}]{minixhofer2024zero}
Benjamin Minixhofer, Edoardo~Maria Ponti, and Ivan Vuli{\'c}. 2024.
\newblock \href {https://arxiv.org/abs/2405.07883} {Zero-shot tokenizer transfer}.
\newblock \emph{ArXiv preprint}, abs/2405.07883.

\bibitem[{Nawrot et~al.(2023)Nawrot, Chorowski, Lancucki, and Ponti}]{nawrot2022efficient}
Piotr Nawrot, Jan Chorowski, Adrian Lancucki, and Edoardo~Maria Ponti. 2023.
\newblock \href {https://doi.org/10.18653/v1/2023.acl-long.353} {Efficient transformers with dynamic token pooling}.
\newblock In \emph{Proceedings of the 61st Annual Meeting of the Association for Computational Linguistics (Volume 1: Long Papers)}, pages 6403--6417, Toronto, Canada. Association for Computational Linguistics.

\bibitem[{Pinter et~al.(2017)Pinter, Guthrie, and Eisenstein}]{pinter2017mimicking}
Yuval Pinter, Robert Guthrie, and Jacob Eisenstein. 2017.
\newblock \href {https://doi.org/10.18653/v1/D17-1010} {Mimicking word embeddings using subword {RNN}s}.
\newblock In \emph{Proceedings of the 2017 Conference on Empirical Methods in Natural Language Processing}, pages 102--112, Copenhagen, Denmark. Association for Computational Linguistics.

\bibitem[{Rust et~al.(2021)Rust, Pfeiffer, Vuli{\'c}, Ruder, and Gurevych}]{rust2020good}
Phillip Rust, Jonas Pfeiffer, Ivan Vuli{\'c}, Sebastian Ruder, and Iryna Gurevych. 2021.
\newblock \href {https://doi.org/10.18653/v1/2021.acl-long.243} {How good is your tokenizer? on the monolingual performance of multilingual language models}.
\newblock In \emph{Proceedings of the 59th Annual Meeting of the Association for Computational Linguistics and the 11th International Joint Conference on Natural Language Processing (Volume 1: Long Papers)}, pages 3118--3135, Online. Association for Computational Linguistics.

\bibitem[{Schick and Sch{\"u}tze(2019)}]{schick2019attentive}
Timo Schick and Hinrich Sch{\"u}tze. 2019.
\newblock \href {https://doi.org/10.18653/v1/N19-1048} {Attentive mimicking: Better word embeddings by attending to informative contexts}.
\newblock In \emph{Proceedings of the 2019 Conference of the North {A}merican Chapter of the Association for Computational Linguistics: Human Language Technologies, Volume 1 (Long and Short Papers)}, pages 489--494, Minneapolis, Minnesota. Association for Computational Linguistics.

\bibitem[{Schick and Sch{\"u}tze(2020)}]{schick2020bertram}
Timo Schick and Hinrich Sch{\"u}tze. 2020.
\newblock \href {https://doi.org/10.18653/v1/2020.acl-main.368} {{BERTRAM}: Improved word embeddings have big impact on contextualized model performance}.
\newblock In \emph{Proceedings of the 58th Annual Meeting of the Association for Computational Linguistics}, pages 3996--4007, Online. Association for Computational Linguistics.

\bibitem[{Schuster and Nakajima(2012)}]{schuster2012japanese}
Mike Schuster and Kaisuke Nakajima. 2012.
\newblock \href {https://doi.org/10.1109/ICASSP.2012.6289079} {{Japanese and korean voice search}}.
\newblock In \emph{2012 IEEE international conference on acoustics, speech and signal processing (ICASSP)}, pages 5149--5152. IEEE.

\bibitem[{Sennrich et~al.(2016)Sennrich, Haddow, and Birch}]{sennrich2015neural}
Rico Sennrich, Barry Haddow, and Alexandra Birch. 2016.
\newblock \href {https://doi.org/10.18653/v1/P16-1162} {Neural machine translation of rare words with subword units}.
\newblock In \emph{Proceedings of the 54th Annual Meeting of the Association for Computational Linguistics (Volume 1: Long Papers)}, pages 1715--1725, Berlin, Germany. Association for Computational Linguistics.

\bibitem[{Singh et~al.(2024)Singh, Romanou, Fourrier, Adelani, Ngui, Vila-Suero, Limkonchotiwat, Marchisio, Leong, Susanto et~al.}]{singh2024global}
Shivalika Singh, Angelika Romanou, Cl{\'e}mentine Fourrier, David~I Adelani, Jian~Gang Ngui, Daniel Vila-Suero, Peerat Limkonchotiwat, Kelly Marchisio, Wei~Qi Leong, Yosephine Susanto, et~al. 2024.
\newblock Global mmlu: Understanding and addressing cultural and linguistic biases in multilingual evaluation.
\newblock \emph{arXiv preprint arXiv:2412.03304}.

\bibitem[{Sun et~al.(2020)Sun, Hashimoto, Yin, Asai, Li, Yu, and Xiong}]{sun2020adv}
Lichao Sun, Kazuma Hashimoto, Wenpeng Yin, Akari Asai, Jia Li, Philip Yu, and Caiming Xiong. 2020.
\newblock \href {https://arxiv.org/abs/2003.04985} {{Adv-bert: Bert is not robust on misspellings! generating nature adversarial samples on bert}}.
\newblock \emph{ArXiv preprint}, abs/2003.04985.

\bibitem[{Tay et~al.(2022)Tay, Tran, Ruder, Gupta, Chung, Bahri, Qin, Baumgartner, Yu, and Metzler}]{tay2021charformer}
Yi~Tay, Vinh~Q. Tran, Sebastian Ruder, Jai~Prakash Gupta, Hyung~Won Chung, Dara Bahri, Zhen Qin, Simon Baumgartner, Cong Yu, and Donald Metzler. 2022.
\newblock \href {https://openreview.net/forum?id=JtBRnrlOEFN} {Charformer: Fast character transformers via gradient-based subword tokenization}.
\newblock In \emph{The Tenth International Conference on Learning Representations, {ICLR} 2022, Virtual Event, April 25-29, 2022}. OpenReview.net.

\bibitem[{Toraman et~al.(2023)Toraman, Yilmaz, {\c{S}}ahinu{\c{c}}, and Ozcelik}]{toraman2023impact}
Cagri Toraman, Eyup~Halit Yilmaz, Furkan {\c{S}}ahinu{\c{c}}, and Oguzhan Ozcelik. 2023.
\newblock Impact of tokenization on language models: An analysis for turkish.
\newblock \emph{ACM Transactions on Asian and Low-Resource Language Information Processing}, 22(4):1--21.

\bibitem[{Touvron et~al.(2023)Touvron, Martin, Stone, Albert, Almahairi, Babaei, Bashlykov, Batra, Bhargava, Bhosale et~al.}]{touvron2023llama}
Hugo Touvron, Louis Martin, Kevin Stone, Peter Albert, Amjad Almahairi, Yasmine Babaei, Nikolay Bashlykov, Soumya Batra, Prajjwal Bhargava, Shruti Bhosale, et~al. 2023.
\newblock \href {https://arxiv.org/abs/2307.09288} {Llama 2: Open foundation and fine-tuned chat models}.
\newblock \emph{ArXiv preprint}, abs/2307.09288.

\bibitem[{Uzan et~al.(2024)Uzan, Schmidt, Tanner, and Pinter}]{uzan2024greed}
Omri Uzan, Craig~W Schmidt, Chris Tanner, and Yuval Pinter. 2024.
\newblock \href {https://arxiv.org/abs/2403.01289} {Greed is all you need: An evaluation of tokenizer inference methods}.
\newblock \emph{ArXiv preprint}, abs/2403.01289.

\bibitem[{Wang et~al.(2021)Wang, Ruder, and Neubig}]{wang2021multi}
Xinyi Wang, Sebastian Ruder, and Graham Neubig. 2021.
\newblock \href {https://doi.org/10.18653/v1/2021.naacl-main.40} {Multi-view subword regularization}.
\newblock In \emph{Proceedings of the 2021 Conference of the North American Chapter of the Association for Computational Linguistics: Human Language Technologies}, pages 473--482, Online. Association for Computational Linguistics.

\bibitem[{Xu et~al.(2023)Xu, Alon, and Neubig}]{xu2023nearest}
Frank~F. Xu, Uri Alon, and Graham Neubig. 2023.
\newblock \href {https://proceedings.mlr.press/v202/xu23a.html} {Why do nearest neighbor language models work?}
\newblock In \emph{International Conference on Machine Learning, {ICML} 2023, 23-29 July 2023, Honolulu, Hawaii, {USA}}, volume 202 of \emph{Proceedings of Machine Learning Research}, pages 38325--38341. {PMLR}.

\bibitem[{Xue et~al.(2022)Xue, Barua, Constant, Al-Rfou, Narang, Kale, Roberts, and Raffel}]{xue2022byt5}
Linting Xue, Aditya Barua, Noah Constant, Rami Al-Rfou, Sharan Narang, Mihir Kale, Adam Roberts, and Colin Raffel. 2022.
\newblock \href {https://doi.org/10.1162/tacl_a_00461} {{B}y{T}5: Towards a token-free future with pre-trained byte-to-byte models}.
\newblock \emph{Transactions of the Association for Computational Linguistics}, 10:291--306.

\bibitem[{Yang et~al.(2018)Yang, Dai, Salakhutdinov, and Cohen}]{yang2018breaking}
Zhilin Yang, Zihang Dai, Ruslan Salakhutdinov, and William~W. Cohen. 2018.
\newblock \href {https://openreview.net/forum?id=HkwZSG-CZ} {Breaking the softmax bottleneck: {A} high-rank {RNN} language model}.
\newblock In \emph{6th International Conference on Learning Representations, {ICLR} 2018, Vancouver, BC, Canada, April 30 - May 3, 2018, Conference Track Proceedings}. OpenReview.net.

\bibitem[{Yu et~al.(2023)Yu, Simig, Flaherty, Aghajanyan, Zettlemoyer, and Lewis}]{yu2024megabyte}
Lili Yu, Daniel Simig, Colin Flaherty, Armen Aghajanyan, Luke Zettlemoyer, and Mike Lewis. 2023.
\newblock \href {http://papers.nips.cc/paper\_files/paper/2023/hash/f8f78f8043f35890181a824e53a57134-Abstract-Conference.html} {{MEGABYTE:} predicting million-byte sequences with multiscale transformers}.
\newblock In \emph{Advances in Neural Information Processing Systems 36: Annual Conference on Neural Information Processing Systems 2023, NeurIPS 2023, New Orleans, LA, USA, December 10 - 16, 2023}.

\end{thebibliography}

\clearpage
\appendix

\section{Dynamic Tokenization Algorithm for Encoder LMs}
\label{sec:bpe_dynamic_tokenization_lm}
Algorithm~\ref{alg:bpe_dynamic_tokenization_lm} shows the simplified version of the byte-pair encoding inspired merging process we use to dynamic tokenize (i.e., compress) a sequence of tokens.
\begin{algorithm*}
\small
\caption{\textit{Dynamic Tokenization for Encoders}}
\label{alg:bpe_dynamic_tokenization_lm}
\begin{algorithmic}[1]

\State \textbf{Input:}  Tokenized batch data \textit{tokenizedBatch} under initial tokenization, $ T_\text{init}\left(\mathcal{D}_{\text{batch}}\right) $; number of merges $ m $
\State \textbf{Output:} Tokenized batch data under a new, dynamically learned tokenization, $ T_\text{new}\left(\mathcal{D}_{\text{batch}}\right)$

\vspace{4mm}
\Procedure{ApplyDynamicTokenization}{\textit{tokenizedBatch}, $ m $}
    \For{$ i \gets 1 $ \textbf{to} $ m $} 
        \State \textit{pairFreqs} $\gets$ ComputePairFreqs(\textit{tokenizedBatch})
        \State \textit{bestPair} $\gets$ GetMostFrequentPair(\textit{pairFreqs}) 
        \State Apply \textit{bestPair} merge rule to \textit{tokenizedBatch} 
    \EndFor
    \State \Return \textit{tokenizedBatch} \Comment{$\mathcal{D}_\text{batch}$ as $T_\text{new}\left(\mathcal{D}_\text{batch}\right)$}
\EndProcedure
\vspace{4mm}
\end{algorithmic}
\end{algorithm*}

\section{Reproducibility Details}\label{sec:reproducibility_details}
A summary of the hyperparameters we used for training and evaluation is shown in Table~\ref{tab:random_seeds} and Table~\ref{tab:hyperparameters}. For decoders, specifically for MMLU, we used the template shown in Figure~\ref{fig:mmlu_template} with a maximum sequence length of 8192. Additionally, both MMLU and MT-Bench were run with \texttt{bfloat16} precision and a batch size of 1 due to computational constraints. For MT-Bench specifically, we set the maximum number of tokens to be generated to 1024. Finally, all the experiments were conducted using an NVIDIA GeForce RTX 4090 GPU with 24 GB of VRAM, powered by the NVIDIA driver version 525.105.17 and CUDA version 12.0.

\begin{table*}[h!]
\centering
\footnotesize
\begin{tabular}{>{\raggedright\arraybackslash}p{3.0cm} >{\raggedright\arraybackslash}p{2.0cm} >{\raggedright\arraybackslash}p{1.5cm} >{\raggedright\arraybackslash}p{1.5cm}}
\toprule
\textbf{Experiment} & \textbf{Python's \texttt{random}} & \textbf{\texttt{torch random}} & \textbf{\texttt{numpy random}} \\ 
\midrule
Encoder experiments & 42 & 42 & 42 \\
\midrule
Decoder experiments on MMLU & 0 & 1234 & 1234 \\
\midrule
Decoder experiments on MT-Bench & 1234 & 1234 & 1234 \\
\bottomrule
\end{tabular}
\caption{Summary of random seeds used across different experiments.}
\label{tab:random_seeds}
\end{table*}

\begin{table*}[h!]
\centering
\footnotesize
\begin{tabular}{>{\raggedright\arraybackslash}p{2.75cm}| >{\raggedright\arraybackslash}p{2.5cm} >{\raggedright\arraybackslash}p{2.5cm} >{\raggedright\arraybackslash}p{3.0cm} >{\raggedright\arraybackslash}p{2.5cm}}
\toprule
 \textbf{Hyperparameter}        & \textbf{XNLI}: Task Adapter with Original Subword Tokenization & \textbf{XNLI}: Joint Task and Dynamic Tokenization Adapter & \textbf{UNER}: Task Adapter with Original Subword Tokenization \&  Joint Task and Dynamic Tokenization Adapter \\ \midrule
\textbf{Matrix Rank $r$}            & 32 & 128& 256 \\ \midrule
\textbf{Scaling Factor $\alpha$}    & 64 & 256 & 512 \\ \midrule
\textbf{Dropout}                      & \multicolumn{3}{c}{0.3} \\ \midrule
\textbf{Epochs}                       & 10 & \{10, 15\} & 15 \\ \midrule
\textbf{Learning Rate}                & $3 \times 10^{-4}$& $1 \times 10^{-4}$& $3 \times 10^{-4}$ \\ \midrule
\textbf{Batch Size}                   & \multicolumn{3}{c}{32} \\ \midrule
\textbf{Optimiser}                    & \multicolumn{3}{c}{AdamW} \\ \midrule
\textbf{Optimiser Parameters}         & $\epsilon = 10^{-8}$, $\beta_1 = 0.9$, $\beta_2 = 0.999$& $\epsilon = 10^{-8}$, $\beta_1 = 0.9$, $\beta_2 = 0.999$& $\epsilon = 10^{-8}$, $\beta_1 = 0.9$, $\beta_2 = 0.999$ \\ \midrule
\textbf{Scheduler}                    & \multicolumn{3}{c}{Linear, no warmup steps}\\ \midrule
\textbf{Max Sequence Length}          & \multicolumn{3}{c}{128} \\ \midrule
\textbf{Weight Decay}          & \multicolumn{3}{c}{0} \\ \midrule
\textbf{Precision}          & \multicolumn{3}{c}{bfloat16} \\ 
\bottomrule
\end{tabular}
\caption{Summary of hyperparameters used for LoRA training.}
\label{tab:hyperparameters}
\end{table*}

\begin{figure*}[!t]

\begin{custombox}{Prompt Template: 5-shot from the same domain as the current question}
\small
\textlangle{}QUESTION\_1\textrangle{}

\textlangle{}ANSWERS\_1\textrangle{}

Answer: \textlangle{}ANSWER\_1\textrangle{}

.

.

.

\textlangle{}QUESTION\_5\textrangle{}

\textlangle{}ANSWERS\_5\textrangle{}

Answer: \textlangle{}ANSWER\_5\textrangle{}

\vspace{4mm}
This question refers to the following information.

\textlangle{}QUESTION\textrangle{}

\textlangle{}ANSWERS\textrangle{}

Answer:''
\end{custombox}
\caption{5-shot prompt template used for MMLU evaluation}
\label{fig:mmlu_template}
\end{figure*}


\section{Vocabulary Expansion for Dynamic Autoregressive Generation}\label{sec:vocabulary_expasion_decoders}
Dynamic tokenization can not be applied in the same way described in Section~\ref{sec:dynamic_tokenization_bpe} for autoregressive generation since it results in an unbounded vocabulary. However, to still achieve some of the benefits of dynamic tokenization, we introduce a method that expands $\vcal_\text{init}$ to a large (but bounded) size for improved inference efficiency in token generation.



Our approach, similar to CoG~\citep{lan2023copyneed} and NEST~\citep{li2024nearest} in its training-free domain adaptation, uses a large token vocabulary instead of a phrase table, reducing the need for billions of phrases. We significantly expand the initial vocabulary to $\vcal_{\text{large}}$, aiming to include more specialized terms and word variations in English. This moves token granularity from subword-level closer to word-level, improving efficiency of the generation process. 
This vocabulary, while static, integrates with the LM using HN-generated embeddings from~\citet{minixhofer2024zero}, providing a similar flexibility to a dynamic approach, without the need for training the embeddings. Although this approach does not currently include dynamic updates of $\vcal_\text{large}$, it sets the foundation for future dynamic vocabulary adjustments. 

Our proposed approach for decoder LMs requires three steps: \emph{(1)} expanding the vocabulary to a large size; \emph{(2)} deciding on a tokenization; \emph{(3)} constructing an approximate nearest neighbor (ANN) index and populating it with token embeddings.

\rparagraph{Expanding the Vocabulary}
In the first step, we aim to expand the initial vocabulary of a decoder LM, $\vcal_\text{init}$, to a significantly larger vocabulary, $\vcallarge$. To achieve this, we can apply one of the widely used subword tokenizers such as BPE, WordPiece or UnigramLM on a large corpus to obtain a vocabulary of $|\vcallarge| - |\vcal_\text{init}|$ tokens.\footnote{In practice, the SentencePiece package~\citep{kudo2018sentencepiece} is often used, particularly for low-resource languages, because it works without the need for pre-tokenized input} In our method, we use BPE algorithm to find $\vcalnew$ and $\mcal_\text{new}$. We then obtain $\vcallarge=\vcal_\text{init}\cup\vcal_\text{new}$.

\rparagraph{Deciding on a Tokenization Function}
Although we have obtained the new merge rules $\mcal_\text{new}$ specific to $\vcalnew$, integrating these with the source tokenizer's existing rules, $\mcal_\text{init}$, is challenging. This is because the original and new merge rules, even if both derived from BPE, were learned independently and are stored sequentially. An example illustrating the challenges associated with merging $\mcal_\text{new}$ and $\mcal_\text{init}$ is presented in Table~\ref{tab:example_merge_tokenizers}.

\begin{table*}[!ht]
\centering
\footnotesize
\begin{tabular}{@{}p{3.5cm}p{3.5cm}p{3.5cm}p{3.5cm}@{}}
\toprule
& \textbf{Tokenizer 1} & \textbf{Tokenizer 2} & \textbf{Merged tokenizer} \\ \midrule
\textbf{Initial Vocabulary} & a, b, c, d, e & a, b, c, d, e & a, b, c, d, e \\ 
\midrule
\textbf{Merge Tables} & & & \\ \midrule
Rule 1 & `a', `b' $\to$ `ab' & `a', `d' $\to$ `ad' & `a', `b' $\to$ `ab' \\
Rule 2 & `ab', `c' $\to$ `abc' & `ad', `e' $\to$ `ade' & `ab', `c' $\to$ `abc' \\
Rule 3 & `d', `e' $\to$ `de' & `b', `c' $\to$ `bc' & `d', `e' $\to$ `de' \\
Rule 4 & - & - & `a', `d' $\to$ `ad' \\
Rule 5 & - & - & `ad', `e' $\to$ `ade' \\
Rule 6 & - & - & `b', `c' $\to$ `bc' \\ \midrule
\textbf{New Vocabulary} & a, b, c, d, e, ab, abc, de & a, b, c, d, e, ad, ade, bc & a, b, c, d, e, ab, abc, de, ad, \textcolor{red}{ade}, bc \\ \midrule
\midrule
 \multicolumn{4}{c}{\textbf{Example tokenize: `ade'}}\\\midrule
Step 1 & [`a', `d', `e'] & [`a', `d', `e'] & [`a', `d', `e'] \\
Step 2 & - & [`ad', `e'] & [`a', `de'] \\
Step 3 & - & [`ade'] & - \\\midrule
\end{tabular}
\caption{Example illustrating how combining merge rules from two BPE tokenizers results in conflicts when tokenizing ``ade''.}
\label{tab:example_merge_tokenizers}
\end{table*}


These challenges highlight the complexity involved in merging tokenizers and the need for a tokenization function that facilitates merging. To address this, we use a Longest-Prefix (LP) tokenization function,\footnote{\citet{uzan2024greed} show that LP greedy tokenization performs on par or better than other tokenization functions.} denoted $T_\text{LP}$, similar to the default method used by WordPiece when the continuation prefix is set to \texttt{blank} (i.e., no character).

\rparagraph{Obtaining Token Embeddings and Index Construction}
Similar to the previous experiments, we use an HN pre-trained on the decoder LM from \citet{minixhofer2024zero} to obtain token embeddings for all the tokens $t \in \vcallarge$, using Equation~\ref{eq:eq_hn_apply_to_vnew} with the expanded vocabulary. The next token is generated as follows:
\begin{description}
    \item[Step 1: Input Tokenization.] Tokenize the textual prompt $\mathbf{x}$: $T_{\text{LP}}(\mathbf{x})$;
    \item[Step 2: Input Embeddings.] Obtain the input embeddings for each token in $T_{\text{LP}}(\mathbf{x})$ using the hypernetwork: $E_{\phi_{\text{in}}}(T_{\text{LP}}(\mathbf{x}))$;
    \item[Step 3: LM Processing.] Forward the tokenized input to the LM;
    \item[Step 4: Output Embeddings.] Obtain the output embeddings for each token in the vocabulary $\vcallarge$: $E_{\phi_{\text{out}}}(\vcallarge)$;
    \item[Step 5: Compute Probability Distribution.] Compute the probability distribution over the vocabulary $\vcallarge$ using the output embeddings and the last hidden state $\mathbf{h}$:
          \[
          \mathbf{p} = \text{softmax}\left(\mathbf{h} \cdot E_{\phi_{\text{out}}}(\vcallarge)^\top\right).
          \]
    \item[Step 6: Sample Next Token.] Sample the next token from the probability distribution $\mathbf{p}$.
\end{description}

The computational bottleneck of this approach lies in Step 5, involving a costly dot product calculation between the last hidden state $\mathbf{h}$ and the output embeddings transposed matrix $E_{\phi_{\text{out}}}(\mathcal{V}_{\text{large}})^\top$, due to the large vocabulary size. To mitigate this, we implement an ANN index, $\mathcal{I}$, allowing us to use $\mathbf{h}$ to efficiently retrieve the $k$ closest tokens, denoted as $\mathcal{I}_k(\mathbf{h})$. This significantly reduces computational overhead by focusing on the ``closest'' tokens during generation, therefore maintaining the LM's parameter count and avoiding excessive scaling of the embedding matrix --- usually seen in multilingual models. This facilitates using large vocabularies without retraining the embedding layer. Figure~\ref{fig:decoder_ANN_dynamic_tk} illustrates the flow for applying dynamic tokenization to decoders with the ANN.

\begin{figure}[!h]
    \centering
    \includegraphics[width=.5\textwidth,trim={50pt 2pt 0 0}, clip]{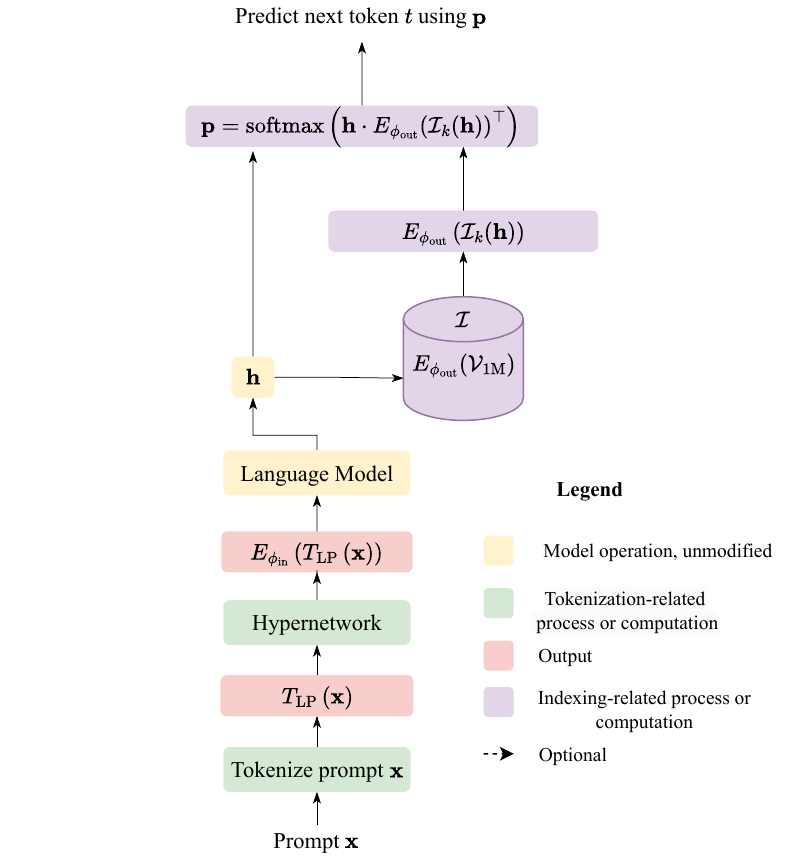}
        \caption{Dynamic tokenization with expanded vocabulary $\vcallarge$ and ANN index $\mathcal{I}$ applied to decoder LMs.}
        \label{fig:decoder_ANN_dynamic_tk}
\end{figure}

\rparagraph{Experimental Setup} In our experiments, we decide to set $\vcallarge$ to one million entries which we denote $\vcalmil$. To obtain this vocabulary $\vcalmil$, we train a BPE tokenizer on the clean English subset of the MADLAD-400 corpus~\citep{kudugunta2024madlad}. Being around twice as large as the Oxford English Dictionary,\footnote{\href{https://www.oed.com}{oed.com}} we expect this vocabulary to contain most English words along other common fragments of text, allowing us to decrease the granularity of the tokens close to word-level on average. This vocabulary is significantly larger than those used in previous works on vocabulary expansion and $\approx32$ times larger than $\vcalinit$.
Following the expansion of the vocabulary, we construct a ScaNN~\citep{guo2020accelerating} index to enable approximate nearest neighbour (ANN) search over HN embeddings. We chose ScaNN due to its good performance on ann-benchmarks.\footnote{\href{https://ann-benchmarks.com/}{ann-benchmarks.com}}
We use this index to retrieve the top 10 closest token embeddings, following the process illustrated in Figure~\ref{fig:decoder_ANN_dynamic_tk}. 

Table~\ref{table:scann_configuration} includes the hyperparameters used for ScaNN index training and inference.

\begin{table}[htbp!] 
\centering
\footnotesize
\centering
\setlength\tabcolsep{5pt} 
\begin{tabularx}{0.8\linewidth}{ll}
\toprule
\multirow{1}{*}{\makecell{\textbf{Attribute}}} & \multicolumn{1}{c}{\makecell{\textbf{Value}}}\\
\midrule
Num. neighbours & 200 \\
Num. leaves & 2000 \\ 
Num. leaves to search & 250 \\
Training Sample Size & 1,000,000 \\
Dim. per block & 3 \\
Anisotropic quantization & 0.2 \\
Reorder & 200 \\
Metric & Dot product \\
\bottomrule
\end{tabularx}
\vspace{0.5mm}
\caption{Configuration details for the ScaNN index.}
\label{table:scann_configuration}
\end{table}

Importantly, when evaluating our approach on MT-Bench, we use both conversation turns rather than just the first turn, as was done in prefilling (see Section~\ref{sec:results_decoders}). This adjustment is made because the current approach with $\vcalmil$ focuses on achieving only (closer to) word-level tokenization, without varying the percentage of sequence length reductions. As a result, determining the number of merges is unnecessary and is in fact difficult to approximate when using two turns, since the second turn contains the model's response to the first turn within its prompt.

\rparagraph{Further findings} When \textsc{Mistral-7B} is evaluated using LP tokenization and HN embeddings with $\vcalmil$ on MT-Bench, we note a decrease in scores of $0.24$ --- compared with the same setting but with $\vcal_\text{init}$. 


Similar to the previous experimental results, the coarser granularity decreases performance, likely due to the quality of the generated HN embeddings. However, the gap between HN and original embeddings is more significant here than in encoders or when applying dynamic tokenization for scoring or prefilling, which could potentially be minimized through n-shot tokenizer transfer, as demonstrated by \citet{minixhofer2024zero}. Additionally, we observe a general trend where performance declines with the use of HN embeddings, worsens further with LP tokenization, and decreases even more with $\vcal_{\text{large}}$, although with the benefit of reduced token sequence length.

\rparagraph{Token repetition penalty improves generations quality} Qualitatively inspecting the MT-Bench generated answers, we observed a token repetition issue in settings with HN embeddings, particularly for prompts from domains requiring creativity (e.g., writing). To address this, we introduced a repetition penalty and top-k sampling with minimum probability threshold. This significantly improved model performance, across all settings using HN embeddings. The token repetition issue may also stem from the \textsc{Mistral-7B} model itself, as multiple reports highlighted similar problems occurring during generation.\footnote{\href{https://huggingface.co/mistralai/Mistral-7B-v0.1/discussions/29}{huggingface.co/mistralai/Mistral-7B-v0.1/discussions/29}} 

\rparagraph{ANN vs Exhaustive Search} Contrary to our expectations, the results indicate that using an ANN index outperforms exhaustive search (setting 9 and 10). This result aligns with findings from other studies, such as those by~\citet{xu2023nearest}, who suggest that the slight ``inaccuracies'' introduced by the ANN index search adds a level of noise or variability, which acts like a regularization technique.

\begin{table*}[!htbp]
\centering
\footnotesize
\setlength\tabcolsep{4.5pt} 
\begin{tabular}{lllll|l}
\toprule
\multirow{2}{*}{\makecell{\textbf{}}} & \multirow{2}{*}{\makecell{\textbf{Tokenization}}} & \multirow{2}{*}{\makecell{\textbf{Embeddings}}} & \multirow{2}{*}{\makecell{\textbf{Vocab.}\\\textbf{Size}}} & \multirow{2}{*}{\makecell{\textbf{$\Delta_\text{Length. (\%)}$}}} & \multirow{2}{*}{\makecell{\textbf{Accuracy}\\\textbf{(\%)}}} \\ \\
\midrule
\rowcolor{gray!20} (1) & original & original & 32k & 0 & \textbf{61.8} \\
(2) & original & HN & 32k & 0 & 58.8 \\
(3) & LP & HN & 32k & \textcolor{JungleGreen}{-1} & 57.8 \\
(4) & LP & HN & 1M & \textcolor{JungleGreen}{-13.6} & 55.9 \\
\bottomrule
\end{tabular}
\vspace{0.5mm}
\caption{
Performance of \textsc{Mistral-7B} on the MMLU English task under different settings. $\Delta_\text{Length. (\%)}$ represents the average decrease in token sequence length for the prompt over the original tokenization. Evaluation was performed under a 5-shot setting with each shot chosen from same domain as the question prompt.}
\label{table:mmlu_results}
\end{table*}

\begin{table*}[!t]
\centering
\footnotesize
\setlength\tabcolsep{4.5pt} 
\begin{tabular}{lllllllc|l}
\toprule
\multirow{2}{*}{\makecell{\textbf{}}} & \multirow{2}{*}{\makecell{\textbf{Tokenization}}} & \multirow{2}{*}{\makecell{\textbf{Embeddings}}} & \multirow{2}{*}{\makecell{\textbf{Vocab.}\\\textbf{Size}}} & \multirow{2}{*}{\makecell{\textbf{Next token}\\\textbf{search}}} & \multirow{2}{*}{\makecell{\textbf{Repetition}\\\textbf{Penalty}}} & \multirow{2}{*}{\makecell{\textbf{Min.}\\\textbf{Prob.}}} & \multirow{2}{*}{\makecell{\textbf{Sample}\\\textbf{from top 10?}}} & \multirow{2}{*}{\makecell{\textbf{Avg}\\\textbf{Score}}} \\ \\
\midrule
\rowcolor{gray!20} (1) & Original & Original & 32k & exhaustive & - & - & \ding{55} & \textbf{7.54} \\
(2) & Original & Original & 32k & exhaustive & 1.1 & 0.05 & \ding{51} & 7.46 \\
\midrule
(3) & Original & HN & 32k & exhaustive & - & - & \ding{55} & 6.84 \\
(4) & Original & HN & 32k & exhaustive & 1.1 & 0.1 & \ding{51} & 7.10 \\
\midrule
(5) & LP & HN & 32k & exhaustive & - & - & \ding{55} & 6.50 \\
(6) & LP & HN & 32k & exhaustive & 1.1 & 0.05 & \ding{51} & \textbf{6.92} \\
\midrule
(7) & LP & HN & 1M & ScaNN index & - & - & \ding{55} & 6.26 \\
(8) & LP & HN & 1M & ScaNN index & 1.1 & 0.05 & \ding{51} &\textbf{ 6.64} \\
\midrule
(9) & LP & HN & 1M & exhaustive & - & - & \ding{55} & 5.24 \\
(10) & LP & HN & 1M & exhaustive & 1.1 & 0.05 & \ding{51} & \textbf{6.53} \\
\bottomrule
\end{tabular}
\vspace{0.5mm}
\caption{
Performance of \textsc{Mistral-7B-Instruct} on MT-Bench English under different settings.} 

\label{table:mtbench_results}
\end{table*}
\section{Hypernetwork Embeddings Caching}\label{sec:emb_cache}
\begin{figure*}[!t] 
\centering    
\includegraphics[width=.8\textwidth]{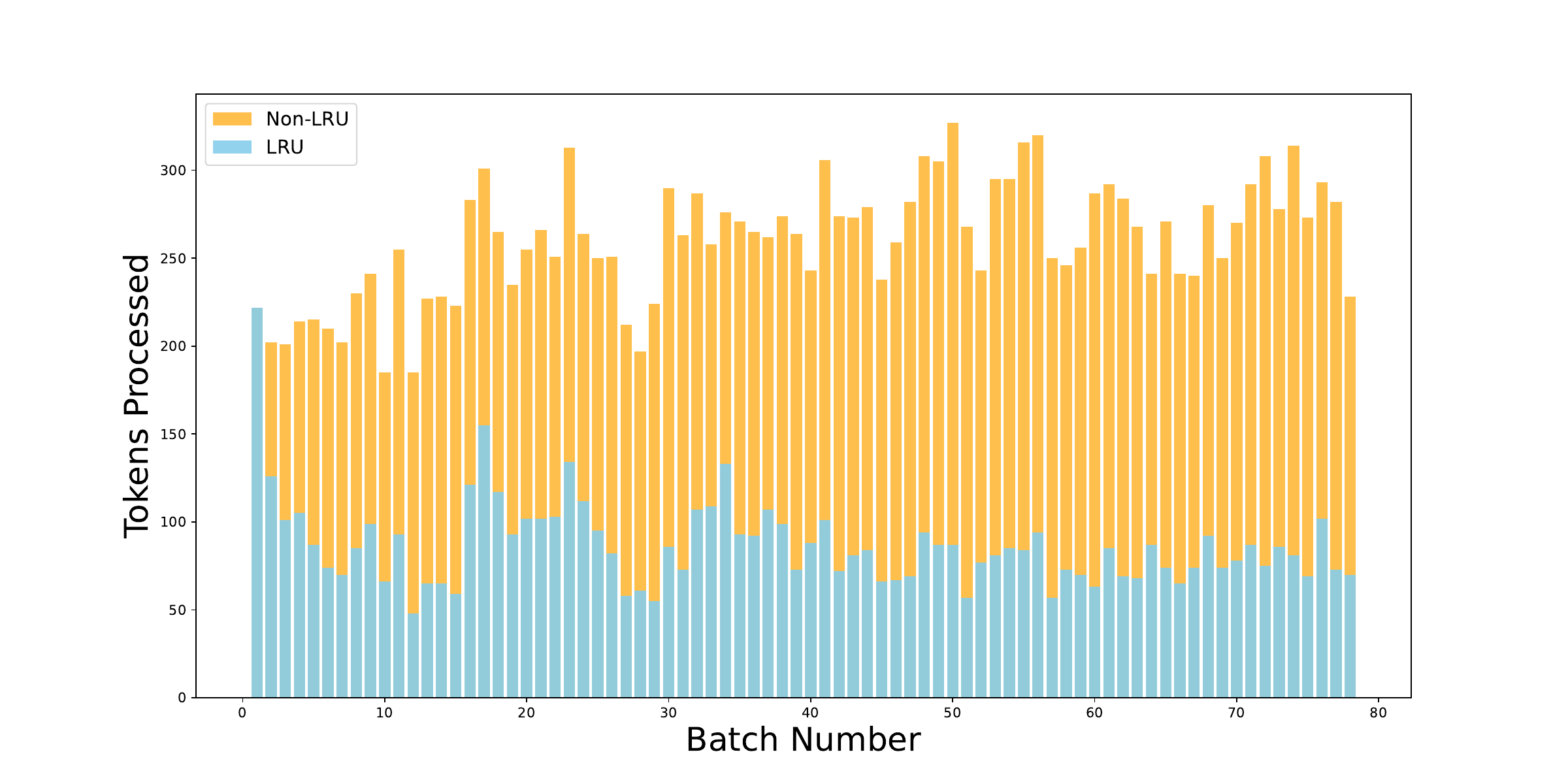}
\caption{Tokens processed by the hypernetwork using an HN-specific LRU cache versus processing all unique tokens without caching. Results obtained on the \texttt{validation} subset of XNLI English.}
\label{fig:hn_processed_tokens}
\end{figure*}

In all our experiments, we implemented a Least-Recently-Used (LRU) cache for storing HN embeddings to enhance efficiency and reduce overhead. This approach was particularly motivated by the frequent repetition of certain tokens across batches in encoder experiments. Common words like ``the'' or ``and'' appear in nearly every utterance, making it practical to cache their embeddings rather than regenerate them for each batch.

\section{Throughput Analysis Results}\label{sec:throughput_analysis}

Table \ref{tab:flops_comparison_main} shows the FLOPs per sample for both the main model and hypernetwork with different sequence reduction percentages on multilingual MMLU. We report results for English, French, German, Spanish, and Portuguese. As shown, the model’s FLOPs decrease almost linearly with sequence length, while the hypernetwork overhead remains negligible --- typically under 3.1\% of total FLOPs. Additionally, the overhead from the dynamic tokenization algorithm is minimal and can be offloaded alongside other data loading logic.

\begin{table*}[!ht]
\centering
\scriptsize
\resizebox{\textwidth}{!}{%
\begin{tabular}{l c c c c c c c c c c c c}
\toprule
\multirow{2}{*}{\textbf{Lng.}} & 
\multirow{2}{*}{\textbf{FLOPs}} & 
\multicolumn{11}{c}{\textbf{Sequence Reduction / FLOPs}} \\
\cmidrule(lr){3-13}
 & & \textbf{0\%} & \textbf{10\%} & \textbf{20\%} & \textbf{30\%} & \textbf{40\%} & \textbf{50\%} & \textbf{60\%} & \textbf{70\%} & \textbf{80\%} & \textbf{90\%} & \textbf{100\%} \\
\midrule
\multirow{4}{*}{\textbf{en}}
   & Model 
     & 10.1T & 10.0T & 9.9T & 9.7T & 9.5T & 9.4T & 9.2T & 9.1T & 8.8T & 8.7T & 8.5T \\
   & Hypernet 
     & 169.3B & 171.3B & 175.0B & 180.2B & 184.7B & 191.0B & 196.8B & 198.7B & 201.5B & 198.0B & 199.8B \\
   & HN FLOPs / total
     & 1.7\% & 1.7\% & 1.7\% & 1.8\% & 1.9\% & 2.0\% & 2.0\% & 2.1\% & 2.2\% & 2.2\% & 2.2\% \\
     \cmidrule(lr){2-13}
   & Seq. Length
     & 682.2 & 672.8 & 667.6 & 655.5 & 640.6 & 631.7 & 619.4 & 614.4 & 598.1 & 586.7 & 578.4 \\
\midrule
\multirow{4}{*}{\textbf{de}} 
   & Model 
     & 15.0T & 14.3T & 13.8T & 13.2T & 12.4T & 11.8T & 11.1T & 10.6T & 9.7T & 9.0T & 8.4T \\
   & Hypernet 
     & 82.9B & 94.6B & 107.8B & 128.9B & 149.3B & 171.1B & 194.7B & 212.2B & 235.5B & 251.3B & 261.4B \\
   & HN FLOPs / total
     & 0.5\% & 0.7\% & 0.8\% & 1.0\% & 1.2\% & 1.4\% & 1.7\% & 1.9\% & 2.2\% & 2.3\% & 3.0\% \\
\cmidrule(lr){2-13}
   & Seq. Length
     & 1015.0 & 963.3 & 937.7 & 882.0 & 825.9 & 782.9 & 748.2 & 711.7 & 668.1 & 608.2 & 571.0 \\
\midrule
\multirow{4}{*}{\textbf{es}} 
   & Model 
     & 14.2T & 13.6T & 13.3T & 12.8T & 12.2T & 11.7T & 11.2T & 10.7T & 10.1T & 9.5T & 9.0T \\
   & Hypernet 
     & 85.2B & 92.0B & 102.1B & 120.4B & 140.5B & 157.7B & 179.1B & 200.5B & 222.9B & 230.6B & 238.1B \\
   & HN FLOPs / total
     & 0.6\% & 0.7\% & 0.8\% & 1.0\% & 1.1\% & 1.3\% & 1.6\% & 1.8\% & 2.1\% & 2.4\% & 2.6\% \\
\cmidrule(lr){2-13}
   & Seq. Length
     & 956.3 & 928.1 & 893.0 & 851.0 & 809.2 & 779.0 & 743.7 & 716.6 & 677.1 & 641.1 & 612.5 \\
\midrule
\multirow{4}{*}{\textbf{fr}}
   & Model 
     & 14.4T & 14.0T & 13.7T & 13.2T & 12.6T & 12.2T & 11.7T & 11.3T & 10.7T & 10.2T & 9.7T \\
   & Hypernet 
     & 91.8B & 99.4B & 109.6B & 128.4B & 146.6B & 163.5B & 183.1B & 200.1B & 223.2B & 230.3B & 238.5B \\
   & HN FLOPs / total
     & 0.6\% & 0.7\% & 0.8\% & 1.0\% & 1.2\% & 1.3\% & 1.5\% & 1.7\% & 2.0\% & 2.2\% & 2.4\% \\
\cmidrule(lr){2-13}
   & Seq. Length
     & 956.7 & 927.9 & 915.3 & 870.7 & 830.9 & 804.6 & 772.6 & 745.9 & 709.1 & 677.8 & 651.9 \\
\midrule
\multirow{4}{*}{\textbf{pt}} 
   & Model 
     & 14.2T & 13.6T & 13.2T & 12.7T & 12.0T & 11.5T & 10.9T & 10.5T & 9.8T & 9.1T & 8.6T \\
   & Hypernet 
     & 84.0B & 89.2B & 101.0B & 118.5B & 131.7B & 154.0B & 175.7B & 196.7B & 218.6B & 229.5B & 237.9B \\
   & HN FLOPs / total
     & 0.6\% & 0.7\% & 0.8\% & 1.0\% & 1.1\% & 1.3\% & 1.6\% & 1.8\% & 2.1\% & 2.3\% & 2.7\% \\
\cmidrule(lr){2-13}
   & Seq. Length
     & 952.3 & 929.7 & 888.2 & 850.4 & 814.5 & 766.0 & 728.9 & 698.3 & 655.0 & 615.3 & 584.8 \\
\bottomrule
\end{tabular}%
} 
\caption{
\textit{FLOPs per sample} estimates for the model and hypernetwork when applying dynamic tokenization with different sequence reductions on multilingual MMLU. The \textit{HN FLOPs / total} row shows the fraction of hypernetwork FLOPs out of total FLOPs. \textit{Seq. Length} represents the average number of tokens per sample.
}
\label{tab:flops_comparison}
\end{table*}

\section{Hypernetwork Training}\label{sec:hypernetwork_training}
We use the pre-trained hypernetworks available via {\faGithub \hspace{0.2mm}} \href{https://github.com/bminixhofer/zett}{github.com/bminixhofer/zett}. Training a hypernetwork for a new ~7B scale model takes $\approx3$ days on a TPUv4-32 pod per \citet{minixhofer2024zero}, which is a small amount of compute compared to pretraining but still substantial. New hypernetworks can be trained via the code at {\faGithub \hspace{0.2mm}} \href{https://github.com/bminixhofer/zett}{github.com/bminixhofer/zett}. By making it easy for the community to train and share hypernetworks, we hope that these libraries facilitate the adoption of our dynamic tokenization method.

\end{document}